\pdfoutput=1

\documentclass[11pt]{article}

\usepackage[final]{acl}

\usepackage{times}
\usepackage{latexsym}
\usepackage{multirow}
\usepackage{makecell}
\usepackage{graphicx,enumitem,amsmath,amssymb,tabularx,booktabs}
\usepackage{subcaption}
\usepackage{arydshln}
\usepackage{subcaption}
\makeatletter
\newcommand\notsotiny{\@setfontsize\notsotiny\@vipt\@viipt}
\makeatother

\newcolumntype{Y}{>{\centering\arraybackslash}X}

\usepackage[T1]{fontenc}

\usepackage[utf8]{inputenc}

\usepackage{microtype}

\usepackage{inconsolata}

\usepackage{graphicx}

\title{Native Design Bias: Studying the Impact of English Nativeness on Language Model Performance}
\author{Manon Reusens$^{1}$, Philipp Borchert$^{1,2}$, Jochen De Weerdt$^{1}$, Bart Baesens$^{1,4}$ \\
         $^1$Research Centre for Information Systems Engineering (LIRIS), KU Leuven \\ %
         $^2$IESEG School of Management, 3 Rue de la Digue, 59000 Lille, France \\ %
         $^4$Department of Decision Analytics and Risk, University of Southampton \\
         \small{\{manon.reusens, philipp.borchert, jochen.deweerdt, bart.baesens\}@kuleuven.be }}

\begin{document}
\maketitle
\begin{abstract}
Large Language Models (LLMs) excel at providing information acquired during pretraining on large-scale corpora and following instructions through user prompts. 
However, recent studies suggest that LLMs exhibit biases favoring Western native English speakers over non-Western native speakers. Given English’s role as a global lingua franca and the diversity of its dialects, we extend this analysis to examine whether non-native English speakers also receive lower-quality or factually incorrect responses more frequently. We compare three groups—Western native, non-Western native, and non-native English speakers---across classification and generation tasks. Our results show that performance discrepancies occur when LLMs are prompted by the different groups for the classification tasks. Generative tasks, in contrast, are largely robust to nativeness bias, likely due to their longer context length and optimization for open-ended responses. Additionally, we find a strong anchoring effect when the model is made aware of the user's nativeness for objective classification tasks, regardless of the correctness of this information. This anchoring effect is a form of cognitive bias shown to be present in LLMs where the model is highly influenced by additional information. Our analysis is based on a newly collected dataset with over 12,000 unique annotations from 124 annotators, including information on their native language and English proficiency.

\end{abstract}

\section{Introduction}
English, as the global lingua franca, is predominant in large-scale text corpora used to train Large Language Models (LLMs)~\citep{ziems-etal-2023-multi,zhang-etal-2023-dont}, including widely used datasets like CommonCrawl. These datasets are primarily tailored to an English-speaking audience located in the United States, and are mainly composed of privileged English dialects from wealthier educated urban zones~\citep{talat-etal-2022-reap,ziems-etal-2023-multi,ryan2024unintended,gururangan-etal-2022-whose}. This biased training dataset composition permeates the LLM, resulting in models tailored to these English dialects~\citep {santy-etal-2023-nlpositionality,hall2022systematic}. This highlights underlying design biases in LLMs, a phenomenon where design choices result in improved downstream performance for specific sub-populations~\citep{santy-etal-2023-nlpositionality}. Consequently, their effectiveness considerably decreases when prompted in other languages or in underrepresented English dialects~\citep{lai-etal-2023-chatgpt,zhang-etal-2023-dont,bang-etal-2023-multitask,ziems-etal-2023-multi,ryan2024unintended}. %

\begin{figure*}[t!]
    \centering
    \begin{subfigure}[b]{0.5\textwidth}
        \centering
        \includegraphics[width=\textwidth]{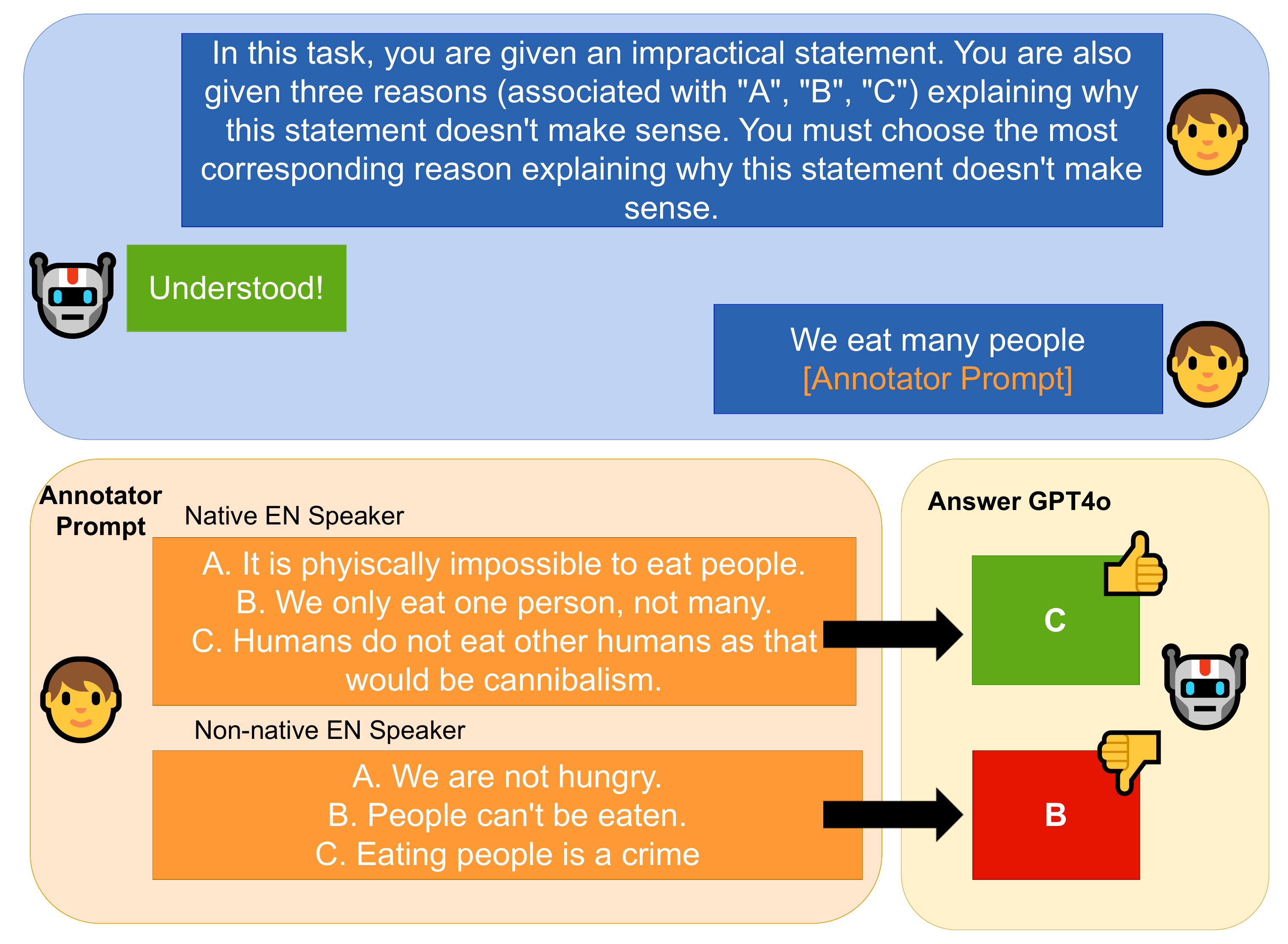}
        \caption{The desired output is C. This is an example prompt of an objective classification task. }
    \end{subfigure}%
    ~ 
    \begin{subfigure}[b]{0.5\textwidth}
        \centering
        \includegraphics[width=\textwidth]{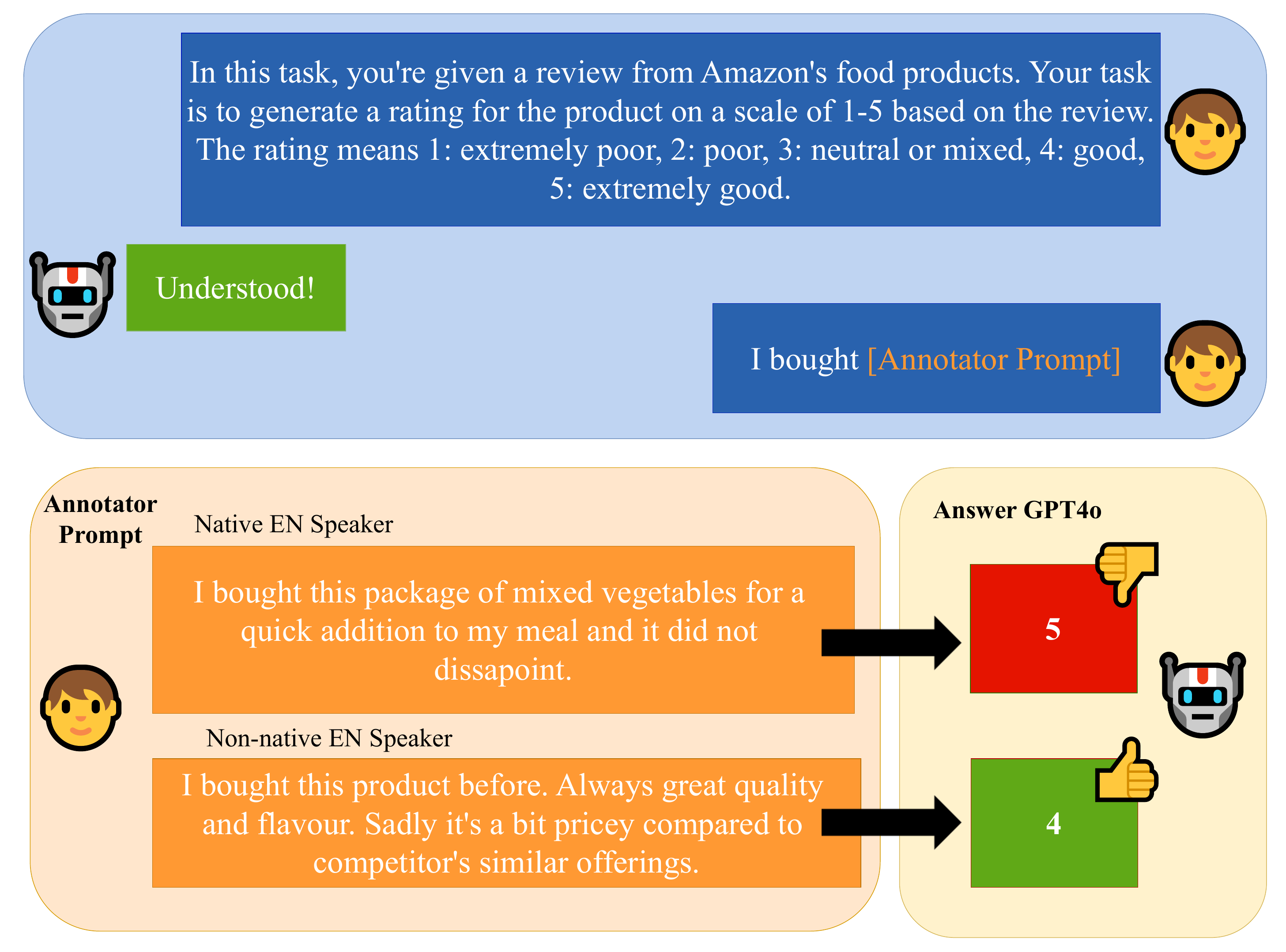}
        \caption{The desired output is 4. This is an example prompt of an subjective classification task. }
    \end{subfigure}
    \caption{Two example prompts of a native and non-native English speaker and the corresponding output given by GPT4o, where \emph{Annotator Prompt} represents the placeholder for the annotations. For the objective task, the model selects the wrong answer for the non-native English speaker, while semantically the same message was conveyed. While Sentence B from the non-native speaker ("People can't be eaten.") may seem different from Sentence A from the native speaker, it is a direct translation from the non-native speaker’s first language and conveys the same meaning from the non-native prompt writer’s perspective. This demonstrates how slight variations in phrasing, common among non-native speakers, can lead to misinterpretations or different model responses, despite semantic equivalence. For the subjective task, we see how the model estimates the native answer to be more positive than actually intended.}
    \label{fig:start_example}
\end{figure*}

\noindent LLMs are highly sensitive to prompt formulations~\cite{beck-etal-2024-sensitivity,chakraborty-etal-2023-zero}. ~\citet{ryan2024unintended} show how models' responses are tailored to Western English dialects, with prompt selection impacting LLMs' preference tuning. Therefore, prompting models in other dialects can result in performance differences due to these design biases. \citet{ziems-etal-2023-multi} even provide a dataset covering multiple English dialects. However, unlike those studies focusing only on English dialects from native English-speaking countries, our research also incorporates participants from countries where English is not an official language. We assess if word sensitivity in prompts disproportionately benefits native English speakers, leading to better model performance. In this case, the model has an inherent native language bias.

\noindent In this paper, we examine performance differences when LLMs are prompted by speakers from three groups: Western native (WN), non-Western native (NWN), and non-native (NN) English speakers. We find performance differences when LLMs are prompted by both NWN and NN versus WN speakers. More specifically, some models generate inaccurate responses for non-native speakers and rate the WN prompts more positively than intended.  We also highlight how LLMs are more robust against this native bias on generative tasks. Moreover, we uncover deeply embedded bias within models towards native speakers for the classification tasks, as explicitly stating that a prompt writer is non-native leads to lower model performance compared to stating that the writer is native regardless of the correctness of this information. We collect a dataset comprising over 12,000 unique prompts from native and non-native English speakers worldwide and demonstrate how different prompt formulations can lead to worse performance despite conveying the same message. An example prompt from our dataset is shown in Figure~\ref{fig:start_example}.  %

\noindent Our contributions are as follows: 1) \textbf{Native bias analysis:} We quantitatively and qualitatively analyze how LLM performance differs between native --- both Western and non-Western--- and non-native English speakers on objective and subjective classification tasks\footnote{By subjective tasks, we mean classification tasks where the correct answer depends on the subjective interpretation as explained in~\citet{beck-etal-2024-sensitivity}}, as well as generative tasks. 2) \textbf{Novel Dataset:} We publish our multilingual instruction-tuning dataset and code used for the experiments\footnote{\url{https://anonymous.4open.science/r/native_en_bias-EDC5/README.md}} containing over 12,000 unique prompts from diverse native and non-native English speakers, with translations into eight languages. 3) \textbf{Innovative Data Collection:} Our large-scale, structured annotation process across various tasks provides a comprehensive view of LLM responses from diverse user groups. 4) \textbf{Novel Experimental Set-up:} We propose a novel design evaluating the impact of informing the model about user nativeness, exploring whether it mitigates bias—an aspect not systematically studied before. %

\section{Related work}

\textbf{Model Positionality and Design Bias.}
Model positionality, coined by \citet{cambo2022model}, refers to the social and cultural position of a model, influenced by the stakeholders involved in its development, such as annotators and developers. This positionality affects the inclusivity of LLMs, as they evolve with certain biases that may disadvantage specific populations~\citep{cambo2022model,santy-etal-2023-nlpositionality}. Design biases arise when researchers make choices that improve model performance for specific sub-populations~\cite{santy-etal-2023-nlpositionality}. A notable example is the overrepresentation of English pretraining corpora, which leads to disproportionate performance improvements in English compared to other languages~\citep{qin-etal-2023-chatgpt,blasi-etal-2022-systematic,joshi-etal-2020-state}.%

\noindent \textbf{Effect of demographic background on LLM performance.}
Recent literature suggests that LLM performance on subjective tasks is influenced by the demographic attributes of the user~\citep{beck-etal-2024-sensitivity,santy-etal-2023-nlpositionality}. Moreover, when assigned a persona, LLMs reveal deep inherent stereotypes against various socio-demographic groups~\citep{cheng-etal-2023-marked,gupta2023bias,deshpande-etal-2023-toxicity}. For example, \citet{gupta2023bias} show how ChatGPT3.5, when asked to solve a math question while adopting the identity of a physically disabled person, generates that it cannot answer the question, as a physically disabled person. Furthermore, \citet{barikeri-etal-2021-redditbias} demonstrate that LLMs can infer demographic attributes from dialog interactions. Additionally, research shows biases in favor of Western populations~\citep{santy-etal-2023-nlpositionality,durmus2023towards}. In model alignment literature, \citet{ryan2024unintended} show this similar bias within preference models and \citet{gururangan-etal-2022-whose} illustrate that even within a Western country like the US, GPT3 prefers the more privileged dialects. Furthermore, \citet{hofmann2024ai} illustrate how models show covert biases towards African American English speakers. Additionally, \citet{kantharuban2024stereotype} show how LLMs express racially stereotypical recommendations regardless of whether the user explicitly or implicitly revealed their identity. Finally, \citet{ziems-etal-2023-multi} have provided a cross-dialectal English dataset for countries with English as an official language.
Building on these findings, we extend the research to include non-native English speakers, who use English dialects influenced by their native languages.
Furthermore, while \citet{gupta2023bias} assign a persona to the model, we analyze performance differences of LLMs both with and without explicitly informing the model about the user's native language and thus with and without assigning a persona to the prompt writer. However, note that models providing different answers based on demographic background is not always problematic as noted in \citet{jin-etal-2024-implicit}.

\section{Methodology} \label{sec:RQs}
Given the sensitivity of LLMs to prompt formulation~\citep{beck-etal-2024-sensitivity,chakraborty-etal-2023-zero}, the diversity of English dialects~\citep{ziems-etal-2023-multi,ryan2024unintended}, and alignment of models towards Western native English speakers~\citep{ryan2024unintended,santy-etal-2023-nlpositionality,gururangan-etal-2022-whose}, we investigate whether LLMs exhibit bias in favor of native English speakers over non-native speakers. More specifically, we aim to answer the following research questions:
 \begin{enumerate}
     \item Do LLMs perform differently when prompted by native vs. non-native English speakers? And is there a performance difference for different groups of native English speakers?
     \item Are certain tasks more prone to performance disparities between native and non-native speakers?
     \item Which tasks, if any, remain robust to these differences?
     \item Are these trends consistent across models, or do they vary by architecture?
     \item Does explicitly providing information about a speaker’s nativeness amplify performance gaps?
\end{enumerate}%

\begin{figure*}
    \centering
    \vspace{-20pt}
    \includegraphics[width=\textwidth]{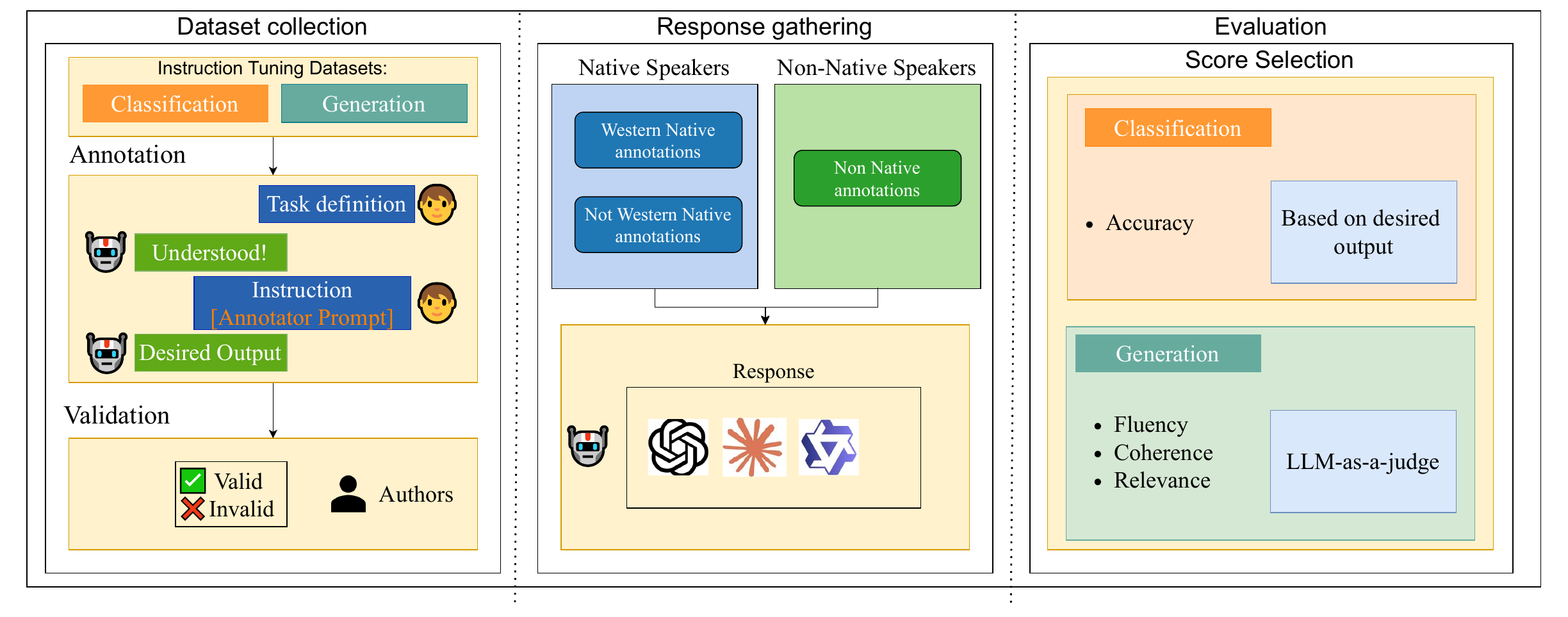}
    \caption{Methodology and experimental setup. The left part shows the data collection steps. After gathering the different datasets, study participants annotated the examples. Then we validated them and used them as input to generate LLM responses. The right part of the figure shows the evaluation phase, where we gathered the respective scores depending on the task.}
    \label{fig:method}
    \vspace{-5pt}
\end{figure*}

 To answer these research questions, we collected a new dataset containing both classification and generation tasks, along with information about the native languages of the annotators, as this is lacking in existing literature. An overview of our methodology and experimental setup is shown in Figure~\ref{fig:method}.
\subsection{Dataset}
Our dataset was constructed including samples from ten diverse task datasets from various natural language instruction tasks\footnote{\url{https://github.com/allenai/natural-instructions}}~\citep{naturalinstructions,supernaturalinstructions}, covering classification (subjective and objective) and generation tasks. These tasks, representing typical LLM interactions, follow a standard instruction pattern and should not inherently favor native speakers. The tasks include paraphrasing, article generation based on a summary or title, sentiment analysis, natural language understanding, multiple-choice answering, and review writing. This last task is the subjective classification task in our experiments. The different tasks provide varying levels of freedom in the dataset annotation tasks. This approach was explicitly chosen to have a range of more and less standardized annotation tasks where the level of freedom in prompt annotations varies depending on the underlying task. This way, our approach provides a comprehensive analysis of model performance.

\noindent From each original dataset, we randomly selected 100 examples, manually ensuring they were correctly annotated and free of offensive language. Additionally, one extra example per dataset served as a tutorial for the annotator to get used to the task. %

 More information about the different tasks included in our dataset can be found in Appendix~\ref{sec:dataset_overview}.

\subsection{Annotations}
We required all annotators to have a minimum English proficiency level equivalent to a high school or university-level proficiency to establish a baseline, ensuring that performance differences stem primarily from dialectal variation rather than overall language proficiency. Each annotator worked on 20 to 240 examples. We gathered them through direct recruitment, opting for an open annotation process rather than an existing annotation platform to ensure high-quality annotations. All annotators were reimbursed at a minimum rate of 12.11 euros per hour. 

In addition to gathering self-reported linguistic data---such as native language, English proficiency, and frequency of English use---we also collected information from native English speakers about how they acquired the language. This allows us to compare three groups: the non-native speakers (NN), Western native speakers (WN), and non-Western native speakers (NWN). The term \emph{Western native} here refers to native English speakers who learned English from native speakers from countries like the UK, US, Australia, or Canada. %

Annotators performed different tasks depending on the assigned datasets. An example annotation is shown in Figure~\ref{fig:start_example}, where a task definition is provided together with an impractical statement. The annotator has to provide the \texttt{[Annotator PROMPT]} based on the task definition and the desired output, which is \emph{C} in this example. We identified the \texttt{[Annotator PROMPT]} per example depending on the dataset. More details about the annotation setup including information about the annotator prompts per dataset can be found in Appendix~\ref{sec:annotation_set-up}.

The authors manually validated the annotations before including them in the final dataset, deeming one invalid if it met any of the following criteria: 1) The response was unrelated to the task, i.e. \textit{"I don't know / understand"}, or a response for a different topic or question. 2) The response contained (part of) the answer. 3) The response did not follow the required format or task definition. 4) The annotator misunderstood the task.
Examples per validation criterion are included in Appendix~\ref{sec:val}.

After validation, we removed instances with more than 50\% rejected annotations to ensure the quality of the dataset. In total, we removed 12 examples entirely and a total of 162 individual annotations. Our final dataset contains 12,519 annotations from 124 annotators.
More information on the dataset statistics can be found in Appendix~\ref{sec:dataset_statistics}\footnote{Due to the nature of the tasks, we did not calculate inter-annotator agreement scores, as annotators were providing prompts, and invalid prompts were filtered out.}.

We thus enforced strict quality control through the data collection phase to mitigate annotator variability through manual validation, removal of low-quality responses, and filtering examples with over 50\% rejected annotations. This ensures that performance differences reflect linguistic or model-driven effects.

\section{Experimental setup}
\subsection{Gathering LLM responses}
Using gathered annotations, we conducted experiments with the chat-versions of well-established LLMs, as these are used in daily life. An overview of the checkpoints per model is shown in Appendix~\ref{sec:checkpoints}. We included GPT3.5\footnote{\url{https://openai.com/index/gpt-3-5-turbo-fine-tuning-and-api-updates/}}, GPT4o\footnote{\url{https://openai.com/index/hello-gpt-4o/}},  Haiku~\citep{anthropic2024claude}, Sonnet~\citep{anthropic2024claude}, using the appropriate APIs, and Qwen1.5 7B\footnote{We ran the experiments for Qwen using A100 GPUs.}~\citep{bai2023qwen} in line with the provided licenses and all consistent with the intended use. This set includes models of varying sizes, different performances, and from different developers, ensuring a diverse representation. Moreover, Qwen, developed by Chinese researchers, provides an interesting comparison in terms of design bias.%

To answer our predefined research questions mentioned in Section~\ref{sec:RQs}, we first ran our experiments for all models without any additional information. Next, to answer the last research question, we provided information about the nativeness of the prompt writer to the LLM. To see whether the LLM entails an inherent bias against native speakers, we included both correct and incorrect information.

\subsection{Evaluation}
To measure the bias within the models, we look into the performance difference between the native and non-native speaking groups. These performance disparities could contribute to allocational harms and representational harms, as defined by~\citet{blodgett-etal-2020-language}. Allocational harms can arise if non-native prompts result in systematically lower-quality responses, potentially affecting users' access to accurate information, career guidance, or educational support. Similarly, representational harms may arise if certain English varieties are implicitly treated as less legitimate, reinforcing linguistic hierarchies and marginalizing speakers of underrepresented dialects. 

\noindent We measure these performance differences across classification tasks and generative tasks. Concretely, native bias measured for the classification tasks is defined as follows:  
\vspace{-5pt}

\vspace{-7pt}
\begin{flalign*}
    \Delta_{\text{native}} &= \phi\left(\mathcal{M}\left(\mathcal{T} \mid x_{\text{native}}\right),\psi\right) \\
    \Delta_{\text{non-native}} &= \phi\left(\mathcal{M}\left(\mathcal{T} \mid x_{\text{non-native}}\right),\psi\right)
\end{flalign*}

\noindent with native bias discriminative = $\Delta_{\text{native}}-\Delta_{\text{non-native}}$, template $\mathcal{T}$, user prompt $x$, model $\mathcal{M}$, accuracy $\phi$, and original ground truth $\psi$.
The native generative bias is defined as follows:
\vspace{-12pt}

\begin{flalign*}
    \Delta_{\text{native}} &= \phi\left(\mathcal{M}\left(\mathcal{T} \mid x_{\text{native}}\right)\right) \\
    \Delta_{\text{non-native}} &= \phi\left(\mathcal{M}\left(\mathcal{T} \mid x_{\text{non-native}}\right)\right)
\end{flalign*}

\noindent with native bias generative = $\Delta_{\text{native}}-\Delta_{\text{non-native}}$,  template $\mathcal{T}$, user prompt $x$, model $\mathcal{M}$, and performance metric $\phi$. The Western native bias can be similarly inferred by splitting the native group into a Western native and non-Western native group.\\

\noindent \textbf{Classification tasks.} When assessing classification tasks, both objective and subjective classification tasks, we focus on the accuracy of the predictions. We only consider classifications as correct if they follow the instructions correctly or if the correct classification can be determined automatically. \\ 

\noindent \textbf{Generative tasks.} In assessing the generative tasks, we include the following metrics: fluency, coherence, and relevance~\citep{Bavaresco2024JUDGE_BENCH}. All metrics were evaluated using a Likert scale: fluency was rated on a 3-point scale. Coherence and relevance were scored on a 5-point scale. Fluency is defined as the quality of the generated text in terms of grammar, spelling, etc. Coherence assesses the collective quality of the sentences. Finally, relevance refers to the inclusion of important content in the generated text. These definitions are based on the ones used in~\citet{Bavaresco2024JUDGE_BENCH}. The prompt templates used are shown in Appendix~\ref{sec:eval}. All results were rescaled to a range of 0 to 1 to ensure clarity. We evaluated the performance of the generative tasks using an LLM-as-a-judge approach, specifically leveraging Llama-3.3-70B-Instruct to assess each prompt’s output according to the three generative metrics mentioned earlier. For transparency, we have also included the exact evaluation prompts in Appendix G. To ensure reliability of the LLM-generated responses, we manually annotated 100 examples and observed a correlation of 81.3\% with the model's evaluations. The Cohen's kappa score is 0.5564. However, given that the generation results are evaluated on a three- and five-point Likert scale, the correlation score is the most informative metric.

\section{Results}
Below, we analyze the results from our experiments answering each of the research questions. Throughout the next paragraphs, we analyze the performance of the native speakers--- consisting of Western native speakers (WN) and  native speakers that are non-Western (NWN)--- and non-native English speakers (NN).
\begin{figure}
    \centering
    \includegraphics[width=\linewidth]{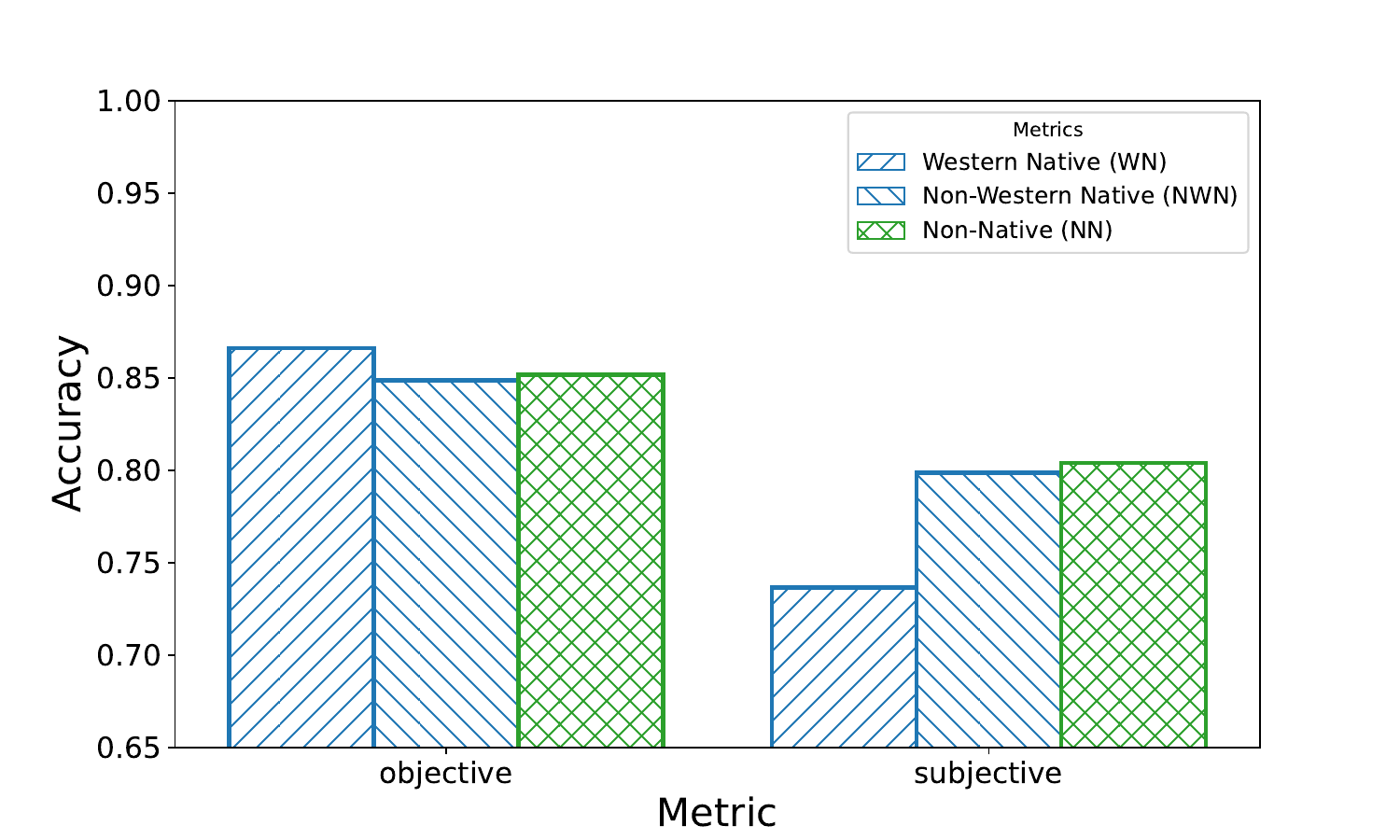}
    \caption{WN is best-performing for objective classification tasks and worst-performing for the subjective classification tasks. The figure shows average model performance per group and task type averaged for all models and runs; y-axis is adjusted to 0.65–1 for clarity.}
    \label{fig:class_results_normal_prompt}
\vspace{-15pt}
\end{figure}

\paragraph{The WN group performs best for the objective classification tasks, outperforming both NWN and NN.}
This is shown in Figure~\ref{fig:class_results_normal_prompt}, where the average performance per group on the objective classification tasks is displayed on the left. WN speakers achieve the highest overall performance in objective classification tasks, reinforcing findings from previous research \cite{hofmann2024ai, ryan2024unintended} that models favor Western privileged dialects. In contrast, NWN and NN English speakers perform similarly, with the NN group slightly outperforming NWN speakers. However, this difference is minimal and not substantial enough to draw strong conclusions. The performance gap between WN and the other groups, however, suggests the advantage of Western dialects. Manual analysis reveals how LLM misclassifications stem from ambiguities in non-native prompts. In Timetravel, less fluent phrasing made incorrect options appear plausible, while in McTaco and TweetQA, non-native formulations led to misinterpretations. This highlights an inherent model bias toward native speakers rather than annotation inconsistencies.

\paragraph{The WN group performs worst for the subjective classification task as models predict their rating more positively than actually intended.} The right part of Figure~\ref{fig:class_results_normal_prompt} shows
this opposite effect for the subjective classification tasks. For these tasks, both the NN and NWN show again similar performance and are now outperforming the WN group.  This finding is remarkable, as it contradicts the results in the subjective classification literature~\citep{santy-etal-2023-nlpositionality,durmus2023towards}. When further analyzing the results, we find that for the Western native English-speaking group, we find that the models often predict the rating more positively than actually intended. While for the NN and NWN groups, GPT4o predicted around 50\% of all wrongly predicted annotations to be more positive than intended, this was around 70\% for the WN English-speaking group for GPT4o indicating cultural differences. Appendix~\ref{app:amazon} includes more information on the different answer distributions per model.

\begin{figure}[t!]
    \centering
    \includegraphics[width=\linewidth]{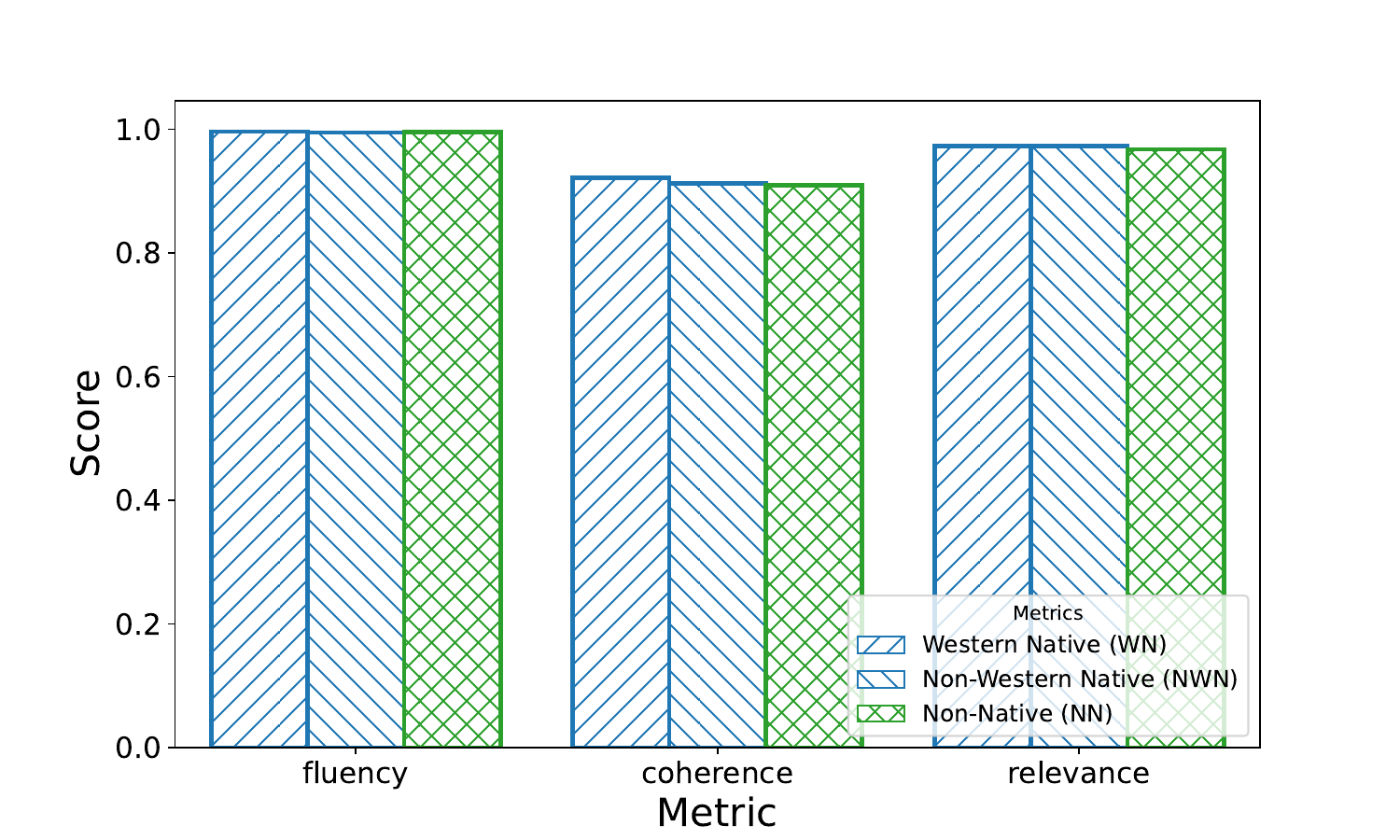}
    \caption{The generative tasks are more robust against native bias. This figure shows the average model performance for the generative tasks per group and metric averaged over the different runs. We rescaled the results so that they range from 0 to 1.}
    \label{fig:gen_results}
\end{figure}

\paragraph{The generation tasks are more robust against (Western) native bias.}
Figure~\ref{fig:gen_results} shows the average performance scores for all models and groups. The figure shows that no clear performance difference exists among the groups compared to the classification results. A slight performance difference favoring the WN group is found for coherence, with the NWN and NN groups performing similarly. Nevertheless, the performance differences are not substantial. Therefore, we conclude that generation tasks are rather robust against (Western) native bias. %
Nevertheless, when zooming in on the results, we find discrepancies depending on the specific task at hand.  These are shown in Appendix~\ref{app:gen_results_overall_per_model}. For two of the datasets, namely Story Cloze and Paraphrase, we find differences in terms of the coherence scores. More specifically, the WN group is here outperforming both the NWN and NN groups. Interestingly, these two tasks also include the smallest written annotations by the prompt writer and generated text by the model. Additionally, when analyzing the CNN DailyMail responses, we find differences in summarization styles among groups. We find how non-native speakers tend to stick closer to the original text when summarizing, while native speakers summarize more freely. Finally, the CODA19 dataset comprises medical articles that utilize specialized medical terminology. Given that most annotators were unfamiliar with this vocabulary, native English speakers (WN and NWN) did not have a specific advantage over non-native speakers. Additionally, research articles are commonly written in English by authors from various backgrounds. Therefore, this specific task might be robust against the native versus non-native preference.

\begin{figure*}
    \centering
    \includegraphics[width=\linewidth]{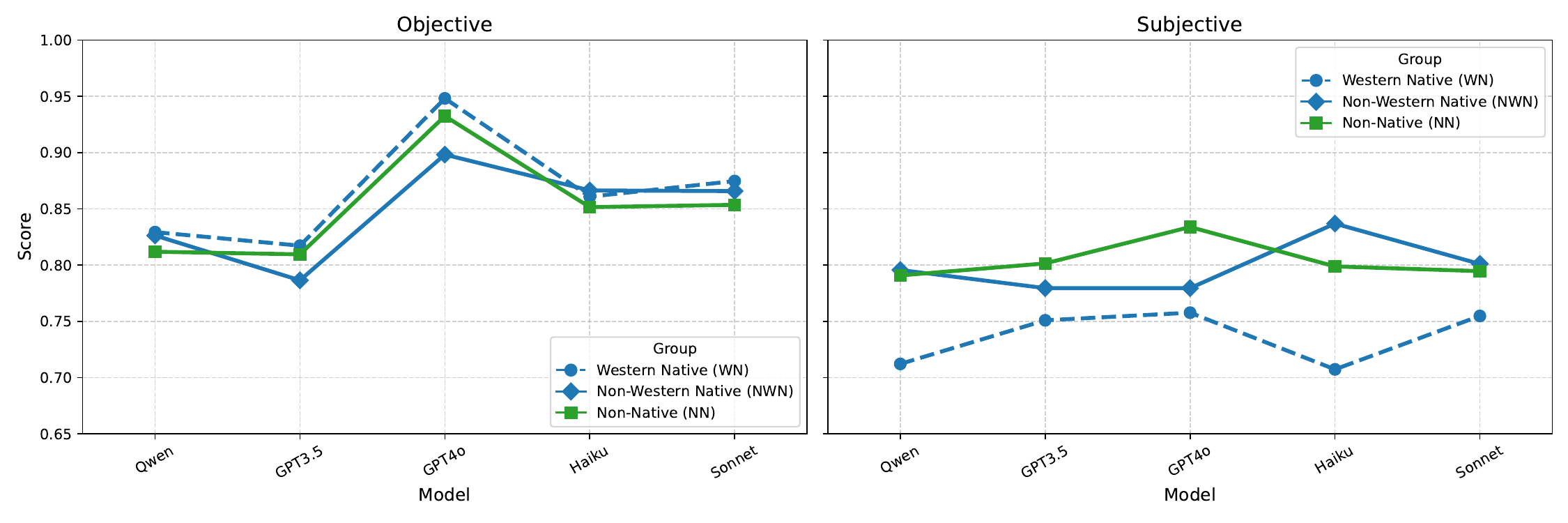}
    \caption{This figure shows the average performance for the different classification tasks per model and group. We see how both GPT models clearly prefer the Western native group, while the other models show similar preference for both native groups for the objective classification task. For the subjective classification tasks, the Western native group is the worst performing group for all models. We adjusted the y-axis to range from 0.65 to 1 for clarity.}
    \label{fig:class_results_obj_subj}
\end{figure*}

\paragraph{(Western) Native bias is model-dependent for the classification tasks.}
Figure~\ref{fig:class_results_obj_subj} illustrates that the preference for WN speakers over NWN speakers in objective classification tasks varies by model. Notably, this trend is pronounced in GPT-3.5 and GPT-4o, while Qwen and Claude models show little to no performance difference between WN and NWN speakers. Interestingly, OpenAI’s models even appear to favor NN speakers over NWN speakers.
 Moreover, it is interesting to see how the Qwen model, developed by Chinese researchers shows almost on par results between both native groups. Additionally, within a model family, the performance disparity increases with model size and overall capability. This aligns with prior research showing a positive correlation between model size and biases, such as gender bias \cite{tal-etal-2022-fewer}. Furthermore, \citet{sclar2023quantifying} demonstrate that prompt sensitivity does not decrease as models scale, suggesting that larger models may reinforce rather than mitigate biases. Also for the subjective classification tasks, the results are strongly model-dependent. However, all models do provide the lowest performance for the WN group. For the generative results, on the other hand, all models show similar trends as is shown in Appendix~\ref{app:gen_results_overall_per_model}.

\begin{figure}[t!]
    \centering
    \includegraphics[width=\linewidth]{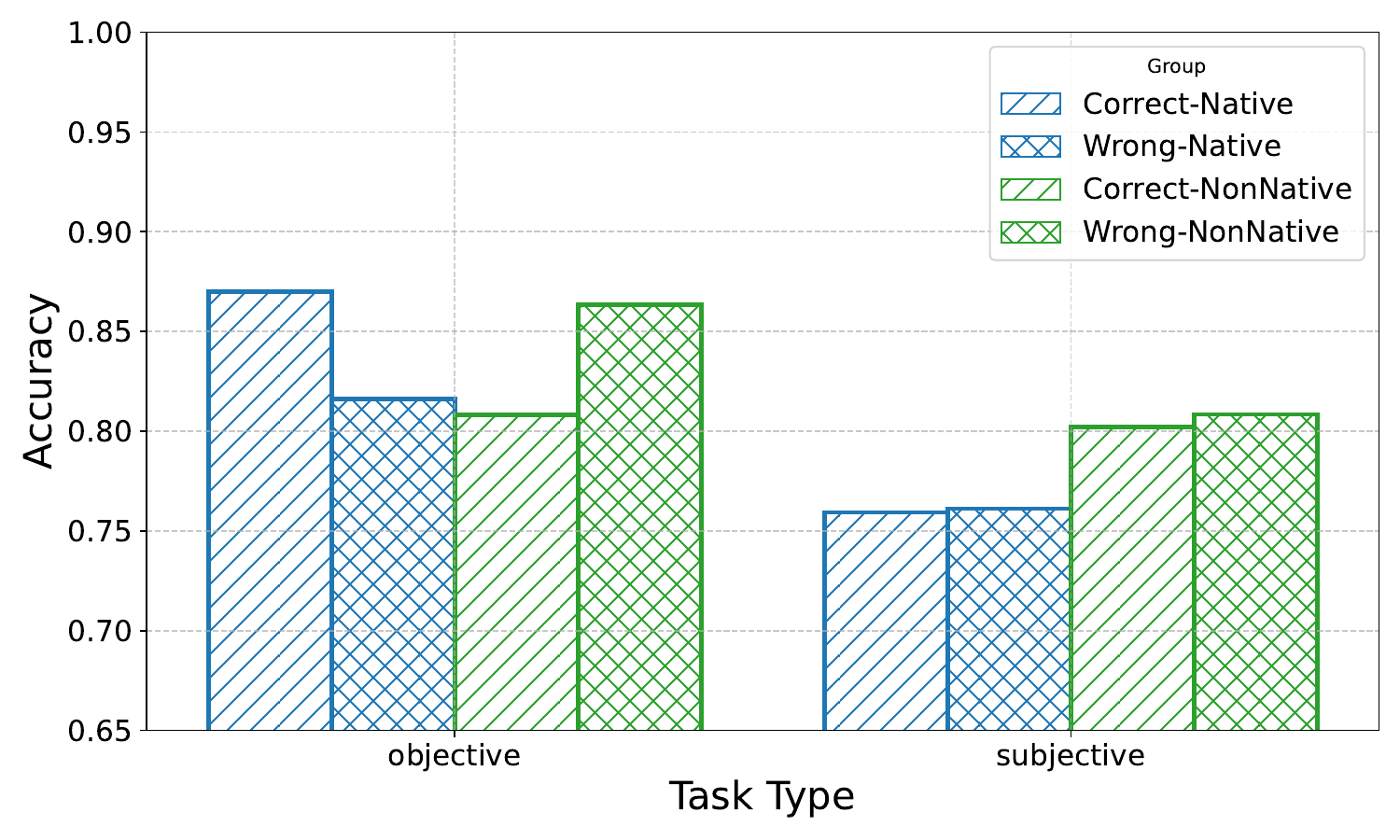}
    \caption{Performance drops when the model is told the prompt writer is non-native rather than native, regardless of the correctness of this information, for objective tasks. Subjective tasks are more robust to this anchoring effect. The figure shows average performance per group and task type based on (in)correct nativeness information averaged over the different models and runs; y-axis is adjusted to 0.65-1 for clarity.}
    \label{fig:class_native_prompt}
\vspace{-20pt}
\end{figure}

\paragraph{Objective classification tasks are largely affected by adding information about the nativeness of the prompt writer.}
Figure~\ref{fig:class_native_prompt} shows the effect of providing the model with (in)correct information about the nativeness of the annotator on model performance. This figure clearly shows how the additional information of the nativeness highly affects the results. Adding correct information about the nativeness results in a clear performance preference for the native group, while adding incorrect information results in a preference for the non-native group.  Moreover, it not only shows how the performance is influenced by this information, but it also reveals deeply embedded bias towards non-native speakers. Adding this information results in a different performance, where the model focuses more on the initial given information than on the prompt itself. This phenomenon is called anchoring.%
 This term is used for human cognitive bias indicating that a person might insufficiently change its estimates away from an initially provided value~\citep{jones2022capturing,tversky1974judgment}. This effect is demonstrated in LLMs by~\citet{jones2022capturing}, who found that code generation models modify their outputs to align with related solutions included in the prompt. Moreover, also  \citet{NGUYEN2024100971} shows how LLM responses are highly influenced by previously given information.  Our results reveal a similar anchoring effect, where the model focuses on the additional information about the nativeness of the prompt writer, regardless of whether or not this information is correct.
This anchoring effect was most clearly present for Sonnet. We find that Sonnet answered several questions in languages other than English, such as Spanish, French, or Indonesian, when responding as if interacting with non-native speakers. This resulted in a clear drop in performance as is also shown in Figure~\ref{fig:class_results_per_model_nativeness} in Appendix~\ref{app:class_results_overall_per_model}. Note that this occurred both for native and non-native speakers. From the other models, we see that Qwen and GPT4o seem to be most robust against this added information. GPT3.5 and Haiku did show performance differences, however, not as pronounced as Sonnet. We manually analyzed examples for GPT3.5 and Haiku to gather more insight into the performance difference. GPT3.5 makes more mistakes when informed about the prompt writer being non-native, due to repetition of the instructions, rather than answering the question. Haiku explains the answers, arguing why one option is better than another, thereby failing to follow the instructions. If both answers are mentioned, we classify the response as inaccurate.

\begin{figure}
    \centering
    \includegraphics[width=\linewidth]{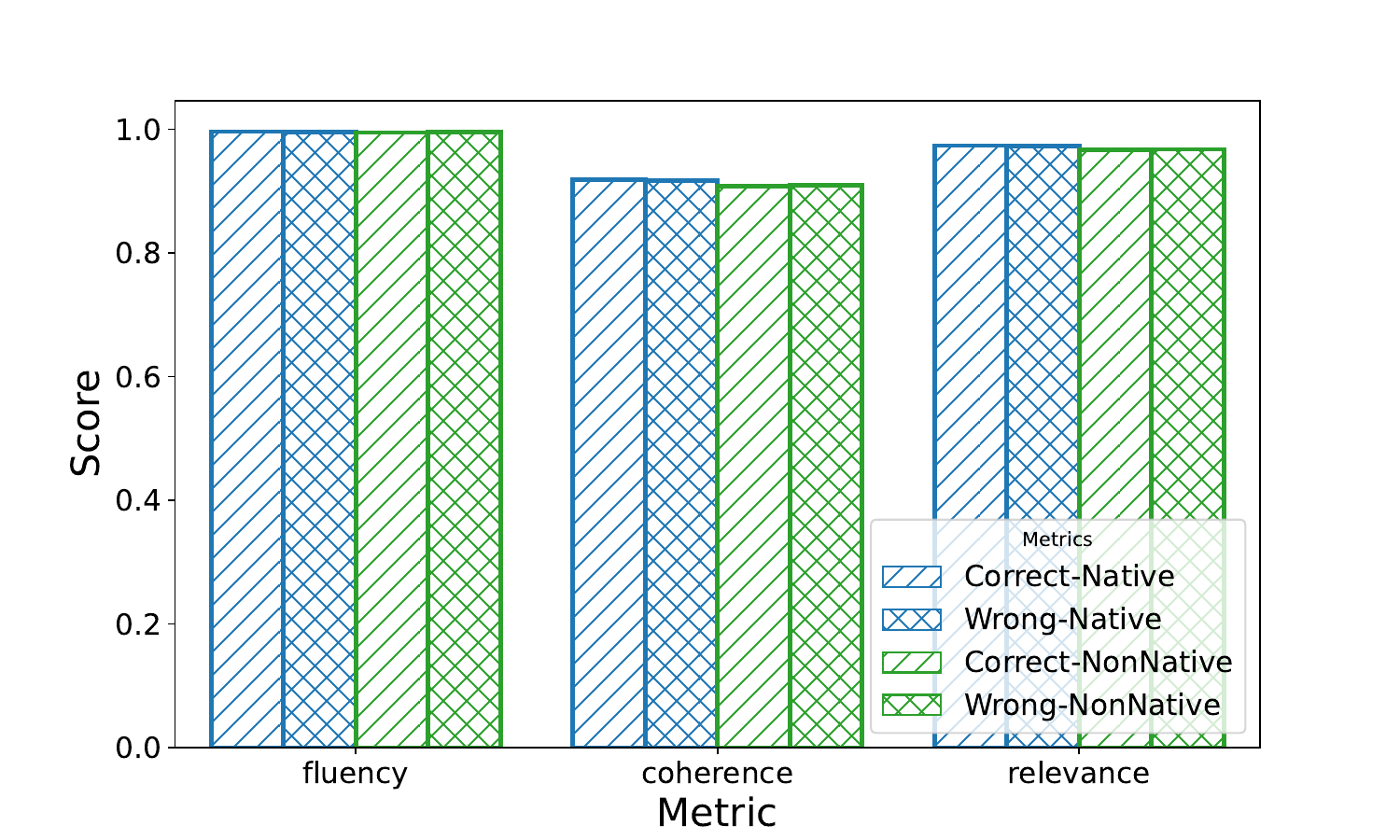}
    \caption{Generative tasks are more robust against the anchoring effect. The figure shows average performance per group and metric averaged over the different models and runs based on (in)correct nativeness information; We rescaled the results so that they range from 0 to 1.}
    \label{fig:gen_resultsnativeness}
\end{figure}

\paragraph{The subjective classification tasks and generation tasks are more robust against this additional information about the prompt writer's nativeness.}
In the subjective classification tasks, we observed only slight performance differences, with the non-native group consistently outperforming the native group. These experiments appear largely unaffected by the addition of information, as the non-native group remains the best-performing regardless of whether accurate or inaccurate details about nativeness are introduced. Also for the generative tasks, the addition of information about the prompt writer's nativeness does not impact performance ranking, as shown in Figure~\ref{fig:gen_resultsnativeness}. All different groups continue to perform similarly, regardless of the additional information provided.%
These findings suggest that models are more robust to nativeness cues in generative  and subjective classification tasks than in objective classification tasks. This is likely due to their primary optimization for generation rather than classification, particularly given that we use the chat-based versions. Additionally, the longer context in both the initial prompt and generated output may reduce the impact of the anchoring effect.

\section{Discussion}
In our experiments, we define native bias as the model's performance disparity when prompted by native versus non-native English speakers. Additionally, we also further split the native speakers into two groups: Western Natives (WN) and Non-Western Natives (NWN). \textbf{In general, we find that there are performance differences when the model is prompted by people from different backgrounds.} 

More specifically, we find an interesting overall preference towards the WN group, where the NWN group is performing similarly as the NN group. This aligns with literature showing how models are tuned towards western native English dialects.  The subjective classification tasks on the other hand, favor the western native group the least, across all different models, contradicting findings from~\cite{santy-etal-2023-nlpositionality,durmus2023towards}. This is explained by the models interpreting the (western) native results more positively than intended.

When analyzing the generative results, we find that this task type is more robust and performing similarly for all evaluated groups. This is probably due to the longer context length in both input and generated output, which seems to help the models to perform similarly across different groups. Additionally, the model checkpoints used were also optimized for these generative tasks rather than classification tasks. 

We show how the performance ranking of the three groups are also model-dependent for the classification tasks. More specifically, the two GPT models are even preferring the NN group over the NWN group on objective classification tasks. The other models show similar performance for both native groups, for the objective classification tasks. Interestingly, Qwen is thus not showing this clear WN preference, but rather a general native preference, similar to Sonnet and Haiku. This is especially interesting given that Qwen is made by researchers that are not based in the US. In literature, studies showing Western native bias have also been conducted on models made by researchers from Western countries~\citep{hofmann2024ai,santy-etal-2023-nlpositionality,durmus2023towards}. However, as was shown by~\citet{buyl2024large}, different models have different ideologies, which in turn influence the different biases entailed in the models. Furthermore, also for the subjective tasks, we see how group preference depends on the model. Nevertheless, all models perform worst for the WN group. The generative tasks on the other hand seem to perform similarly across all models.

Finally, we show how a strong \textbf{anchoring effect} occurs when the model is made aware of the nativeness of the prompt writer for the objective classification tasks. The bias is so deeply engraved that informing the models about the nativeness of both groups results in a preference towards the group that was indicated as native, regardless of the correctness of this information, being led by this additional information rather than by the prompt itself. This anchoring effect has been shown to exist in LLMs for a wide range of applications~\citep{jones2022capturing,NGUYEN2024100971,echterhoff-etal-2024-cognitive}. \citet{echterhoff-etal-2024-cognitive} analyze the existence of cognitive bias in decision-making with LLMs, while~\citet{NGUYEN2024100971} focus on using LLMs for financial forecasting. Finally,~\citet{jones2022capturing} focus in a case study on code generation. Our analyses show the existence of this anchoring effect for the objective classification tasks observing differences across models. GPT4o appears most resistant to this anchoring effect, while Sonnet on the other hand even changes the language of the response based on this anchor. \citet{echterhoff-etal-2024-cognitive} similarly find how GPT4 seems less prone to the anchoring effect than GPT3.5. \citet{NGUYEN2024100971} on the other hand, find the opposite for financial forecasting. Nevertheless, given that LLMs are not optimized for this task, this could also affect the conclusions. In our experiments, we also find that the anchoring effect is not clearly present for the generative results, probably due to the optimization of these models towards generative tasks compared to classification tasks, given that we used chat-versions.

\section{Conclusion}
In this work, we analyze bias in LLMs towards native English speakers. We analyze if models perform better for native compared to non-native English speakers and whether the models are even further tuned towards Western native English speakers. We find that there are performance differences between native and non-native prompts. More specifically, models are most accurate for the Western-native English speakers on objective classification tasks. A slightly lower performance is shown for the NWN group compared to the NN, nevertheless, we show that this is mostly model-dependent. Both GPT models seem even to prefer NN over NWN, while the other models in our analysis show similar performance for both native groups. Furthermore, we find a strong anchoring effect when information about the user's nativeness is added for objective classification tasks. Generative tasks seem to be in general more robust against this native bias, probably due to the longer context length and the optimization of the used models towards these generative tasks. For our experiments, we used a newly collected dataset consisting of over 12,000 unique prompts from a diverse set of annotators.

\section{Limitations}
Our dataset contained a very diverse set of annotators. Nevertheless, it would be interesting to have more study participants for every sub-population, such that general findings at sub-population level could be made as well. The annotations in the dataset are done by annotators from different groups. However, there is an imbalance in number of native English speakers compared to the number of non-native English speakers. Furthermore, our experiments contained mostly annotators having a self-reported level of English of C1 and C2. It would be very interesting to analyze the effects on the performance of LLMs when prompted by people having different levels of English as this will probably also be impactful. Additionally, our results were only gathered for five different models. It would be insightful to extend this analysis to more models, as every model is trained differently and therefore these design choices might lead to different biases within the model.
An important limitation of using LLMs and especially the closed-source variant thereof, is the lack of reproducibility of the results. We make available a multilingual dataset, however, have only analyzed the English answers. We leave the analysis of bias in the multilingual dataset for future research. Additionally, some of the datasets contain a Western focus in terms of the topics that are discussed (CNN Dailymail, TweetQA, and McTaco). While other datasets, like the Amazon Food reviews, are based on user-generated content and may be less culturally specific, we recognize that the overall selection may still reflect Western contexts.  Finally, we acknowledge how the LLM-as-a-judge implementation for gathering generative results might be suboptimal to human annotators due to model-specific biases. Therefore, we chose a different LLM than the ones we will evaluate to serve as a judge to avoid self-preference bias and we manually validated a sample. To further assess the reliability we have included both the correlation and cohen's kappa score. Given the high score for both metrics between the manual annotations and the LLM annotations, we assume that the LLM annotations are representative. %

\section{Ethical considerations}
We included human annotators in this study. All annotators were paid for the provided annotations and the annotations were done on a voluntary base. Moreover, our paper shows some of the consequences of unfair design choices when developing models. We think this work is important to highlight the necessity of taking into account multiple English dialects, as these models should work equally well for everyone. In this paper, we focus on the English language. We wanted to point out that even in English, this problem of not having enough diversified training data might also result in performance differences among certain populations. However, this does not mean that other languages do not require the same attention.

\bibliography{bibliography,anthology}

@inproceedings{lai-etal-2023-chatgpt,
    title = "{C}hat{GPT} Beyond {E}nglish: Towards a Comprehensive Evaluation of Large Language Models in Multilingual Learning",
    author = "Lai, Viet  and
      Ngo, Nghia  and
      Pouran Ben Veyseh, Amir  and
      Man, Hieu  and
      Dernoncourt, Franck  and
      Bui, Trung  and
      Nguyen, Thien",
    editor = "Bouamor, Houda  and
      Pino, Juan  and
      Bali, Kalika",
    booktitle = "Findings of the Association for Computational Linguistics: EMNLP 2023",
    month = dec,
    year = "2023",
    address = "Singapore",
    publisher = "Association for Computational Linguistics",
    url = "https://aclanthology.org/2023.findings-emnlp.878",
    doi = "10.18653/v1/2023.findings-emnlp.878",
    pages = "13171--13189",
}

@inproceedings{zhang-etal-2023-dont,
    title = "Don{'}t Trust {C}hat{GPT} when your Question is not in {E}nglish: A Study of Multilingual Abilities and Types of {LLM}s",
    author = "Zhang, Xiang  and
      Li, Senyu  and
      Hauer, Bradley  and
      Shi, Ning  and
      Kondrak, Grzegorz",
    editor = "Bouamor, Houda  and
      Pino, Juan  and
      Bali, Kalika",
    booktitle = "Proceedings of the 2023 Conference on Empirical Methods in Natural Language Processing",
    month = dec,
    year = "2023",
    address = "Singapore",
    publisher = "Association for Computational Linguistics",
    url = "https://aclanthology.org/2023.emnlp-main.491",
    doi = "10.18653/v1/2023.emnlp-main.491",
    pages = "7915--7927",
}

@inproceedings{qin-etal-2023-chatgpt,
    title = "Is {C}hat{GPT} a General-Purpose Natural Language Processing Task Solver?",
    author = "Qin, Chengwei  and
      Zhang, Aston  and
      Zhang, Zhuosheng  and
      Chen, Jiaao  and
      Yasunaga, Michihiro  and
      Yang, Diyi",
    editor = "Bouamor, Houda  and
      Pino, Juan  and
      Bali, Kalika",
    booktitle = "Proceedings of the 2023 Conference on Empirical Methods in Natural Language Processing",
    month = dec,
    year = "2023",
    address = "Singapore",
    publisher = "Association for Computational Linguistics",
    url = "https://aclanthology.org/2023.emnlp-main.85",
    doi = "10.18653/v1/2023.emnlp-main.85",
    pages = "1339--1384",
}

@inproceedings{beck-etal-2024-sensitivity,
    title = "Sensitivity, Performance, Robustness: Deconstructing the Effect of Sociodemographic Prompting",
    author = "Beck, Tilman  and
      Schuff, Hendrik  and
      Lauscher, Anne  and
      Gurevych, Iryna",
    editor = "Graham, Yvette  and
      Purver, Matthew",
    booktitle = "Proceedings of the 18th Conference of the European Chapter of the Association for Computational Linguistics (Volume 1: Long Papers)",
    month = mar,
    year = "2024",
    address = "St. Julian{'}s, Malta",
    publisher = "Association for Computational Linguistics",
    url = "https://aclanthology.org/2024.eacl-long.159",
    pages = "2589--2615",
}

@inproceedings{cambo2022model,
  title={Model positionality and computational reflexivity: Promoting reflexivity in data science},
  author={Cambo, Scott Allen and Gergle, Darren},
  booktitle={Proceedings of the 2022 CHI Conference on Human Factors in Computing Systems},
  pages={1--19},
  year={2022}
}

@inproceedings{santy-etal-2023-nlpositionality,
    title = "{NLP}ositionality: Characterizing Design Biases of Datasets and Models",
    author = "Santy, Sebastin  and
      Liang, Jenny  and
      Le Bras, Ronan  and
      Reinecke, Katharina  and
      Sap, Maarten",
    editor = "Rogers, Anna  and
      Boyd-Graber, Jordan  and
      Okazaki, Naoaki",
    booktitle = "Proceedings of the 61st Annual Meeting of the Association for Computational Linguistics (Volume 1: Long Papers)",
    month = jul,
    year = "2023",
    address = "Toronto, Canada",
    publisher = "Association for Computational Linguistics",
    url = "https://aclanthology.org/2023.acl-long.505",
    doi = "10.18653/v1/2023.acl-long.505",
    pages = "9080--9102",
}

@inproceedings{cheng-etal-2023-marked,
    title = "Marked Personas: Using Natural Language Prompts to Measure Stereotypes in Language Models",
    author = "Cheng, Myra  and
      Durmus, Esin  and
      Jurafsky, Dan",
    editor = "Rogers, Anna  and
      Boyd-Graber, Jordan  and
      Okazaki, Naoaki",
    booktitle = "Proceedings of the 61st Annual Meeting of the Association for Computational Linguistics (Volume 1: Long Papers)",
    month = jul,
    year = "2023",
    address = "Toronto, Canada",
    publisher = "Association for Computational Linguistics",
    url = "https://aclanthology.org/2023.acl-long.84",
    doi = "10.18653/v1/2023.acl-long.84",
    pages = "1504--1532",
}

@article{gupta2023bias,
  title={Bias runs deep: Implicit reasoning biases in persona-assigned llms},
  author={Gupta, Shashank and Shrivastava, Vaishnavi and Deshpande, Ameet and Kalyan, Ashwin and Clark, Peter and Sabharwal, Ashish and Khot, Tushar},
  journal={arXiv preprint arXiv:2311.04892},
  year={2023}
}

@inproceedings{deshpande-etal-2023-toxicity,
    title = "Toxicity in chatgpt: Analyzing persona-assigned language models",
    author = "Deshpande, Ameet  and
      Murahari, Vishvak  and
      Rajpurohit, Tanmay  and
      Kalyan, Ashwin  and
      Narasimhan, Karthik",
    editor = "Bouamor, Houda  and
      Pino, Juan  and
      Bali, Kalika",
    booktitle = "Findings of the Association for Computational Linguistics: EMNLP 2023",
    month = dec,
    year = "2023",
    address = "Singapore",
    publisher = "Association for Computational Linguistics",
    url = "https://aclanthology.org/2023.findings-emnlp.88",
    doi = "10.18653/v1/2023.findings-emnlp.88",
    pages = "1236--1270",
}

@inproceedings{naturalinstructions,
  title={Cross-task generalization via natural language crowdsourcing instructions},
  author={Mishra, Swaroop and Khashabi, Daniel and Baral, Chitta and Hajishirzi, Hannaneh},
  booktitle={ACL},
  year={2022}
}

@inproceedings{supernaturalinstructions,
  title={Super-NaturalInstructions:Generalization via Declarative Instructions on 1600+ Tasks},
  author={Wang, Yizhong and Mishra, Swaroop and Alipoormolabashi, Pegah and Kordi, Yeganeh and Mirzaei, Amirreza and Arunkumar, Anjana and Ashok, Arjun and Dhanasekaran, Arut Selvan and Naik, Atharva and Stap, David and others},
  booktitle={EMNLP},
  year={2022}
}

@misc{bai2023qwen,
      title={Qwen Technical Report}, 
      author={Jinze Bai and Shuai Bai and Yunfei Chu and Zeyu Cui and Kai Dang and Xiaodong Deng and Yang Fan and Wenbin Ge and Yu Han and Fei Huang and Binyuan Hui and Luo Ji and Mei Li and Junyang Lin and Runji Lin and Dayiheng Liu and Gao Liu and Chengqiang Lu and Keming Lu and Jianxin Ma and Rui Men and Xingzhang Ren and Xuancheng Ren and Chuanqi Tan and Sinan Tan and Jianhong Tu and Peng Wang and Shijie Wang and Wei Wang and Shengguang Wu and Benfeng Xu and Jin Xu and An Yang and Hao Yang and Jian Yang and Shusheng Yang and Yang Yao and Bowen Yu and Hongyi Yuan and Zheng Yuan and Jianwei Zhang and Xingxuan Zhang and Yichang Zhang and Zhenru Zhang and Chang Zhou and Jingren Zhou and Xiaohuan Zhou and Tianhang Zhu},
      year={2023},
      eprint={2309.16609},
      archivePrefix={arXiv},
      primaryClass={cs.CL}
}

@article{anthropic2024claude,
  title={The claude 3 model family: Opus, sonnet, haiku},
  author={Anthropic, AI},
  journal={Claude-3 Model Card},
  year={2024}
}

@inproceedings{bang-etal-2023-multitask,
    title = "A Multitask, Multilingual, Multimodal Evaluation of {C}hat{GPT} on Reasoning, Hallucination, and Interactivity",
    author = "Bang, Yejin  and
      Cahyawijaya, Samuel  and
      Lee, Nayeon  and
      Dai, Wenliang  and
      Su, Dan  and
      Wilie, Bryan  and
      Lovenia, Holy  and
      Ji, Ziwei  and
      Yu, Tiezheng  and
      Chung, Willy  and
      Do, Quyet V.  and
      Xu, Yan  and
      Fung, Pascale",
    editor = "Park, Jong C.  and
      Arase, Yuki  and
      Hu, Baotian  and
      Lu, Wei  and
      Wijaya, Derry  and
      Purwarianti, Ayu  and
      Krisnadhi, Adila Alfa",
    booktitle = "Proceedings of the 13th International Joint Conference on Natural Language Processing and the 3rd Conference of the Asia-Pacific Chapter of the Association for Computational Linguistics (Volume 1: Long Papers)",
    month = nov,
    year = "2023",
    address = "Nusa Dua, Bali",
    publisher = "Association for Computational Linguistics",
    url = "https://aclanthology.org/2023.ijcnlp-main.45",
    doi = "10.18653/v1/2023.ijcnlp-main.45",
    pages = "675--718",
}

@inproceedings{chakraborty-etal-2023-zero,
    title = "Zero-shot Approach to Overcome Perturbation Sensitivity of Prompts",
    author = "Chakraborty, Mohna  and
      Kulkarni, Adithya  and
      Li, Qi",
    editor = "Rogers, Anna  and
      Boyd-Graber, Jordan  and
      Okazaki, Naoaki",
    booktitle = "Proceedings of the 61st Annual Meeting of the Association for Computational Linguistics (Volume 1: Long Papers)",
    month = jul,
    year = "2023",
    address = "Toronto, Canada",
    publisher = "Association for Computational Linguistics",
    url = "https://aclanthology.org/2023.acl-long.313",
    doi = "10.18653/v1/2023.acl-long.313",
    pages = "5698--5711",
}

@article{hall2022systematic,
  title={A systematic study of bias amplification},
  author={Hall, Melissa and van der Maaten, Laurens and Gustafson, Laura and Jones, Maxwell and Adcock, Aaron},
  journal={arXiv preprint arXiv:2201.11706},
  year={2022}
}

@article{ryan2024unintended,
  title={Unintended Impacts of LLM Alignment on Global Representation},
  author={Ryan, Michael J and Held, William and Yang, Diyi},
  journal={arXiv preprint arXiv:2402.15018},
  year={2024}
}

@inproceedings{ziems-etal-2023-multi,
    title = "Multi-{VALUE}: A Framework for Cross-Dialectal {E}nglish {NLP}",
    author = "Ziems, Caleb  and
      Held, William  and
      Yang, Jingfeng  and
      Dhamala, Jwala  and
      Gupta, Rahul  and
      Yang, Diyi",
    editor = "Rogers, Anna  and
      Boyd-Graber, Jordan  and
      Okazaki, Naoaki",
    booktitle = "Proceedings of the 61st Annual Meeting of the Association for Computational Linguistics (Volume 1: Long Papers)",
    month = jul,
    year = "2023",
    address = "Toronto, Canada",
    publisher = "Association for Computational Linguistics",
    url = "https://aclanthology.org/2023.acl-long.44",
    doi = "10.18653/v1/2023.acl-long.44",
    pages = "744--768",
}

@article{durmus2023towards,
  title={Towards measuring the representation of subjective global opinions in language models},
  author={Durmus, Esin and Nyugen, Karina and Liao, Thomas I and Schiefer, Nicholas and Askell, Amanda and Bakhtin, Anton and Chen, Carol and Hatfield-Dodds, Zac and Hernandez, Danny and Joseph, Nicholas and others},
  journal={arXiv preprint arXiv:2306.16388},
  year={2023}
}

@article{jones2022capturing,
  title={Capturing failures of large language models via human cognitive biases},
  author={Jones, Erik and Steinhardt, Jacob},
  journal={Advances in Neural Information Processing Systems},
  volume={35},
  pages={11785--11799},
  year={2022}
}

@article{tversky1974judgment,
  title={Judgment under Uncertainty: Heuristics and Biases: Biases in judgments reveal some heuristics of thinking under uncertainty.},
  author={Tversky, Amos and Kahneman, Daniel},
  journal={science},
  volume={185},
  number={4157},
  pages={1124--1131},
  year={1974},
  publisher={American association for the advancement of science}
}

@article{sclar2023quantifying,
  title={Quantifying Language Models' Sensitivity to Spurious Features in Prompt Design or: How I learned to start worrying about prompt formatting},
  author={Sclar, Melanie and Choi, Yejin and Tsvetkov, Yulia and Suhr, Alane},
  journal={arXiv preprint arXiv:2310.11324},
  year={2023}
}

@article{hofmann2024ai,
  title={AI generates covertly racist decisions about people based on their dialect},
  author={Hofmann, Valentin and Kalluri, Pratyusha Ria and Jurafsky, Dan and King, Sharese},
  journal={Nature},
  volume={633},
  number={8028},
  pages={147--154},
  year={2024},
  publisher={Nature Publishing Group UK London}
}

@inproceedings{jin-etal-2024-implicit,
    title = "Implicit Personalization in Language Models: A Systematic Study",
    author = {Jin, Zhijing  and
      Heil, Nils  and
      Liu, Jiarui  and
      Dhuliawala, Shehzaad  and
      Qi, Yahang  and
      Sch{\"o}lkopf, Bernhard  and
      Mihalcea, Rada  and
      Sachan, Mrinmaya},
    editor = "Al-Onaizan, Yaser  and
      Bansal, Mohit  and
      Chen, Yun-Nung",
    booktitle = "Findings of the Association for Computational Linguistics: EMNLP 2024",
    month = nov,
    year = "2024",
    address = "Miami, Florida, USA",
    publisher = "Association for Computational Linguistics",
    url = "https://aclanthology.org/2024.findings-emnlp.717/",
    doi = "10.18653/v1/2024.findings-emnlp.717",
    pages = "12309--12325"
}

@article{NGUYEN2024100971,
title = {Human bias in AI models? Anchoring effects and mitigation strategies in large language models},
journal = {Journal of Behavioral and Experimental Finance},
volume = {43},
pages = {100971},
year = {2024},
issn = {2214-6350},
doi = {https://doi.org/10.1016/j.jbef.2024.100971},
url = {https://www.sciencedirect.com/science/article/pii/S2214635024000868},
author = {Jeremy K. Nguyen}
}

@inproceedings{Bavaresco2024JUDGE_BENCH,
  title={LLMs instead of Human Judges? A Large Scale Empirical Study across 20 NLP Evaluation Tasks},
  author={Anna Bavaresco and Raffaella Bernardi and Leonardo Bertolazzi and Desmond Elliott and Raquel Fernández and Albert Gatt and E. Ghaleb and Mario Giulianelli and Michael Hanna and Alexander Koller and André F. T. Martins and Philipp Mondorf and Vera Neplenbroek and Sandro Pezzelle and Barbara Plank and David Schlangen and Alessandro Suglia and Aditya K Surikuchi and Ece Takmaz and Alberto Testoni},
  year={2024},
  url={https://arxiv.org/abs/2406.18403}
}

@article{kantharuban2024stereotype,
  title={Stereotype or personalization? user identity biases chatbot recommendations},
  author={Kantharuban, Anjali and Milbauer, Jeremiah and Strubell, Emma and Neubig, Graham},
  journal={arXiv preprint arXiv:2410.05613},
  year={2024}
}

@inproceedings{echterhoff-etal-2024-cognitive,
    title = "Cognitive Bias in Decision-Making with {LLM}s",
    author = "Echterhoff, Jessica Maria  and
      Liu, Yao  and
      Alessa, Abeer  and
      McAuley, Julian  and
      He, Zexue",
    editor = "Al-Onaizan, Yaser  and
      Bansal, Mohit  and
      Chen, Yun-Nung",
    booktitle = "Findings of the Association for Computational Linguistics: EMNLP 2024",
    month = nov,
    year = "2024",
    address = "Miami, Florida, USA",
    publisher = "Association for Computational Linguistics",
    url = "https://aclanthology.org/2024.findings-emnlp.739/",
    doi = "10.18653/v1/2024.findings-emnlp.739",
    pages = "12640--12653"
}

@article{buyl2024large,
  title={Large language models reflect the ideology of their creators},
  author={Buyl, Maarten and Rogiers, Alexander and Noels, Sander and Bied, Guillaume and Dominguez-Catena, Iris and Heiter, Edith and Johary, Iman and Mara, Alexandru-Cristian and Romero, Rapha{\"e}l and Lijffijt, Jefrey and others},
  journal={arXiv preprint arXiv:2410.18417},
  year={2024}
}

\appendix
\section{Dataset overview}
\label{sec:dataset_overview}
We used the datasets as they were assembled by~\citet{naturalinstructions} and \citet{supernaturalinstructions}. Table~\ref{tab:selected_datasets} shows an overview of the selected datasets, together with their task ID in the original instructions dataset.
The task definition given in the table is the one we used when prompting the models. For CNN Dailymail and CODA19, this differs from the original task definition in the dataset because we flipped the task. Instead of letting our annotators write the article, we asked them to write the summary or title respectively. Datasets Abductivenli, Timetravel, Amazonfood, McTaco, TweetQA, and Commonsense are thus classification tasks, while datasets StoryCloze, CNN Dailymail, CODA19, and Paraphrase are generation tasks.

\begin{table*}[]
\tiny
    \centering
    \begin{tabularx}{\textwidth}{ll!{\vrule}X}
    \toprule
         \textbf{Task ID} & \textbf{Name} & \textbf{Task Definition} \\ \midrule
         task069 & Abductivenli & In this task, you will be shown a short story with a beginning,  two potential middles, and an ending. Your job is to choose the middle statement that makes the story  coherent / plausible by writing "1" or "2" in the output.  If both sentences are plausible, pick the  one that makes most sense. \\
         task105 & Story Cloze & In this task, you're given four sentences of a story written in natural language. Your job is to complete the end part of the story by predicting the appropriate last sentence which is coherent with the given sentences.\\
         task065 & Timetravel & In this task, you are given a short story consisting of exactly 5 sentences where the second sentence is missing. You are given two options and you need to select the one that best connects the first sentence with the rest of the story. Indicate your answer by 'Option 1' if the first option is correct, otherwise 'Option 2'. The incorrect option will change the subsequent storyline, so that at least one of the three subsequent sentences is no longer consistent with the story. \\
         task588 & Amazonfood rating & In this task, you're given a review from Amazon's food products. Your task is to generate a rating for the product on a scale of 1-5 based on the review. The rating means 1: extremely poor, 2: poor, 3: neutral or mixed, 4: good, 5: extremely good. \\
         task020 & Mctaco & The answer will be 'yes' if the provided sentence contains an explicit mention that answers the given question. Otherwise, the answer should be 'no'. Instances where the answer is implied from the sentence using "instinct" or "common sense" (as opposed to being written explicitly in the sentence) should be labeled as 'no'. \\
         task241 & TweetQA & In this task, you are given a context tweet, a question and the corresponding answer of the given question. Your task is to classify this question-answer pair into two categories: (1) "yes" if the given answer is right for question, and (2) "no" if the given answer is wrong for question. \\
         task1553 & CNN Dailymail & In this task, you are given highlights ,i.e., a short summary, in a couple of sentences, of news articles and you need to generate the news article with a maximum length of 2 paragraphs. \\
         task1161 & CODA19 & In this task, you're given a title from a research paper and your task is to generate a paragraph for the research paper based on the given title. Under 10 lines is a good paragraph length.  \\
         task177 & Paraphrase & This is a paraphrasing task. In this task, you're given a sentence and your task is to generate another sentence which express same meaning as the input using different words. \\
         task295 & Commonsense & In this task, you are given an impractical statement. You are also given three reasons (associated with "A", "B", "C") explaining why this statement doesn't make sense. You must choose the most corresponding reason explaining why this statement doesn't make sense. \\ \bottomrule
    \end{tabularx}
    \caption{Overview of the different datasets used for the experiments in this paper.}
    \label{tab:selected_datasets}
\end{table*}

\section{Annotation set-up}
\label{sec:annotation_set-up}
\begin{figure}[t!]
    \centering
    \includegraphics[width=0.45\textwidth]{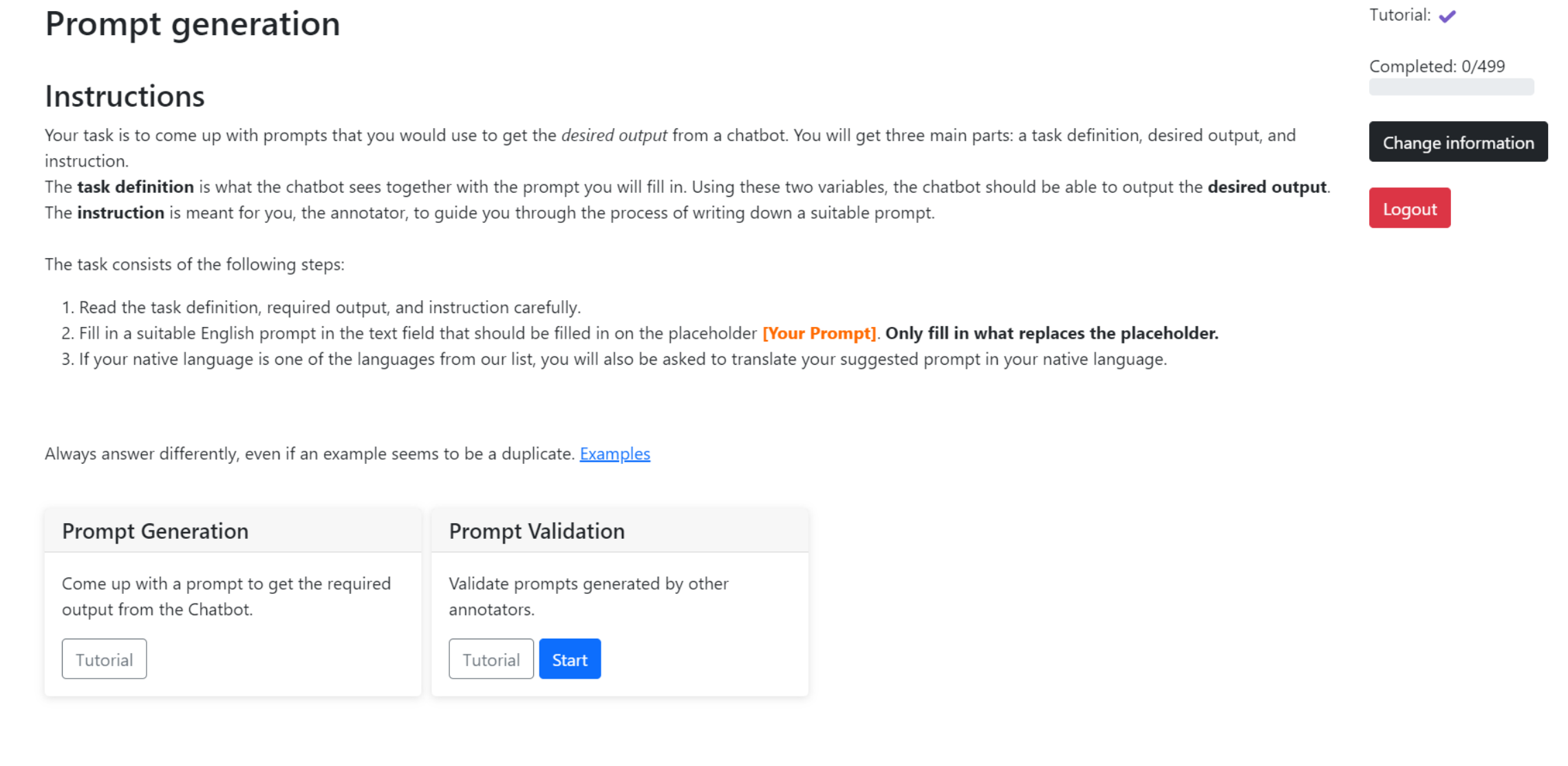}
    \caption{Screenshot of the landing page of the annotation platform.}
    \label{fig:ann_platform}
\end{figure}

\begin{figure}[t!]
    \centering
    \includegraphics[width=0.45\textwidth]{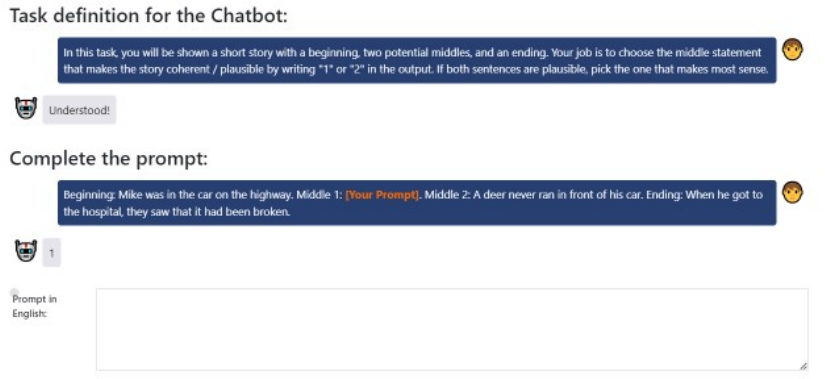}
    \caption{An annotation example of the Abductivenli dataset.}
    \label{fig:annotation_example}
\end{figure}

We have set up an annotation platform to gather the annotations. The annotators first get information about the task. They will get a task definition, a prompt where part of the answer is marked out with the placeholder [YOUR PROMPT], and the desired output of the LLM. The annotators should complete the prompt such that the desired output would be generated by the LLMs. Figure~\ref{fig:ann_platform} shows a screenshot of the landing page of the annotation platform together with annotation instructions. An example of an annotation that had to be annotated is shown in Figure~\ref{fig:annotation_example}. An example of the different \texttt{[Annotator PROMPT]} per dataset is shown in Table~\ref{tab:ann_prompts}. We have anonymized all annotations by only providing the self-reported linguistic information in the dataset along with the user ID number. 

\begin{table*}[]
\footnotesize
    \centering
    \begin{tabularx}{\textwidth}{l!{\vrule}X}
    \toprule
         \textbf{Dataset} & \textbf{Example Prompt}\\ \midrule
         Abductivenli & Beginning: Mike was in the car on the highway. Middle 1 : \texttt{[Annotator Prompt]}. Middle 2: A deer never ran in front of his car. Ending: When he got to the hospital, they saw that it had been broken \\
         Story Cloze & Sentence1: \texttt{[Annotator Prompt]} Sentence2: Suddenly, there was an announcement. Sentence3: The school was on a lockdown. Sentence4: The kids sat quietly, and waited. \\
         Timetravel & Sentence 1: Little Charlie and his dad were painting the garage. Sentence 3: His dad turned around and started to laugh Sentence 4:  Charlie had paint on him from head to toe Sentence 5:  His dad rinsed him off with water from the hose Option 1: \texttt{[Annotator Prompt]} Option 2: Charlie had some trouble controlling the brush. \\
         AmazonFood rating & This is \texttt{[Annotator Prompt]} \\
         McTaco & Sentence: The legitimization of gambling led to its increased legalization across the US. Question: \texttt{[Annotator Prompt]} \\
         TweetQA & Context: Praying for everyone here in Vegas. I witnessed the most unimaginable event tonight. We are okay. Others arent. Please pray. –Jake Owen (@jakeowen) October 2, 2017 Question: \texttt{[Annotator Prompt]} Answer: people were not okay \\
         CNN DailyMail & \texttt{[Annotator Prompt]} \\
         CODA19 & \texttt{[Annotator Prompt]} \\
         Paraphrase & \texttt{[Annotator Prompt]} \\
         Commonsense & I walk under the park. \texttt{[Annotator Prompt]} \\ \bottomrule
    \end{tabularx}
    \caption{Example of a prompt to annotate per dataset. \texttt{[Annotator Prompt]} indicates where the prompt of that the annotator should come up with, should fit in the text.}
    \label{tab:ann_prompts}
\end{table*}

\section{Annotation validation}
\label{sec:val}
Examples for each of the criteria of an invalid annotation are shown in Table~\ref{tab:invalid_criteria}.

\begin{table*}[]
\tiny
    \centering
    \begin{tabularx}{\textwidth}{X!{\vrule}lXl}
    \toprule
         \textbf{Criteria} & \textbf{Dataset} & \textbf{Example} & \textbf{Desired Answer} \\ \midrule
          The response is unrelated to the task or it includes a response for a different topic or question & TweetQA &  Context: I lost the role in 50 Shades of Grey so you won't be hearing from me for awhile— Lena Dunham (@lenadunham) September 2, 2013 Question: which countries are next to France? Answer: liverpool and everybody.& no \\
          The response contains (part of) the answer. & Amazonfood & These are Amazon fish fingers, 5 stars from me - extremely good! & 5 \\
         The response does not follow the required format or task definition. & TweetQA & Context: Kasich's daughter on his dance moves: "You're not going to go on 'Dancing with the Stars'" \#KasichFamily CNN Politics (@CNNPolitics) April 12, 2016 Question: no, as he is terrible at dancing Answer: dozen & no \\
         The person misunderstood the task. & Commonsense & He is wearing a green car choose an alphabet rating for this sentence, "A" for unreasonable meaning, otherwise "B" &  A \\ \bottomrule
    \end{tabularx}
    \caption{Examples for the criteria of an invalid annotation.}
    \label{tab:invalid_criteria}
\end{table*}

For the annotations that did not follow the required format, we tried to change it into the correct format without changing the content of the prompt, if possible (i.e. removing \textit{Question:} ). If this was not possible, the annotation was rejected.

\section{Dataset Statistics -Annotations}
\label{sec:dataset_statistics}

The native-bias dataset consists of 12,519 annotations from 124 annotators. Our dataset initially contained 1,000 different examples. After deleting the examples that were not validly annotated by at least 50 \% of annotators, we retained 988 examples for 10 different tasks.

The annotators have varying native languages as shown in Table~\ref{tab:annotators}. The languages are shown in isocode format. Moreover, per native language, we have also included the average validation rate, that is the amount of annotations per person that were valid over the total number of annotated examples.
\begin{table}[]
\centering
    \resizebox{!}{2.3cm}{\begin{tabularx}{\textwidth}{l!{\vrule}lXl}
    \toprule
        \makecell{\textbf{Native} \\ \textbf{language}} & \makecell{\textbf{Number of} \\ \textbf{annotators}} & \textbf{Languages} & \textbf{Validation rate}   \\ \midrule
        Other & 36 &  BG, SL, RU, SW, ML, HU, FA, VI, BE,  EL, TN, ID,PL, MR, TR,  PT, T, RO,    FIL,  UR, SQ & 0.83 \\ \midrule
        NL & 23 &  & 0.80 \\ \midrule
        EN & 28 &  & 0.83 \\ \midrule
        ZH & 11 &  & 0.82  \\ \midrule
        EN, other & 9 & PA, JA, SW, UR, VI, MR, EL & 0.86  \\ \midrule
        EN, ZH & 1 &  & 0.88  \\ \midrule
        ES & 5 &  & 0.77 \\ \midrule
        FR & 4 &  & 0.94\\ \midrule
        IT & 3 &  & 0.94  \\ \midrule
        HI & 2 &  & 0.93 \\ \midrule
        AR & 1 &  & 0.94   \\ \midrule
        ES, Other & 1 & CA & 0.84  \\ \bottomrule
    \end{tabularx}}
    \caption{Overview of the native languages of the annotators and the validation rate per native language.}
    \label{tab:annotators}
\end{table}

Table~\ref{tab:annotators_per_group} shows an overview of the number of annotators per group and set-id. All annotators were given sets of examples that had to be annotated. Every example has a unique set-id.

\begin{table}[]
\scriptsize
\centering
    \resizebox{!}{1.9cm}{\begin{tabularx}{0.5\textwidth}{l!{\vrule}ll!{\vrule}ll!{\vrule}l}
    \toprule
        \multirow{2}{*}{\makecell{Set ids}} & \multicolumn{2}{l}{Native or not} & \multicolumn{2}{l}{\makecell{Western native or not}}  & \multirow{2}{*}{Total} \\ \cmidrule(lr){2-3} \cmidrule(lr){4-5}
         & Native & \makecell{Non-native} & Western & \makecell{Not Western} & \\ \midrule
        10 & 7 & 16 & 5 & 18 & 23  \\
        20 & 7 & 12 & 4 & 15 & 19 \\
        30 & 7 & 10 & 4 & 13 & 17  \\
        40 & 4 & 8 & 3 & 9 & 12  \\
        50 & 4 & 9 & 2 & 11 & 13  \\
        60 & 5 & 14 & 3 & 16 & 19 \\
        70 & 5 & 11 & 4 & 12 & 16  \\
        80 & 3 & 10 & 3 & 10 & 13  \\
        90 & 4 & 10 & 4 & 10 & 14 \\
        100 & 6 & 5 & 4 & 7 & 11 \\ \bottomrule
    \end{tabularx}}
    \caption{Overview of the number of annotators per group and set.}
    \label{tab:annotators_per_group}
\end{table}

Furthermore, the annotators have reported their level of English proficiency and the frequency of which English was spoken. We provide this information for the non-native speakers in Tables~\ref{tab:enprof} and \ref{tab:enfreq}.
\begin{table}[]
\centering
\small
    \begin{tabularx}{\linewidth}{ll}
    \toprule
        \textbf{English proficiency level} & \makecell{\textbf{Number of non-native}\\ \textbf{ annotators}}  \\ \midrule
        C2 & 31 \\ \midrule
        C1 & 41 \\ \midrule
        B2  & 13  \\ \midrule
        B1 & 1   \\ \bottomrule
    \end{tabularx}
    \caption{Overview of the self-reported English proficiency of the non-native annotators.}
    \label{tab:enprof}
\end{table}

\begin{table}[]
\centering
\small
    \begin{tabularx}{\linewidth}{ll}
    \toprule
        \textbf{English usage frequency} & \makecell{\textbf{Number of non-native}\\ \textbf{ annotators}}  \\ \midrule
        Daily & 60 \\ \midrule
        A few times per week & 21 \\ \midrule
        Once a week  & 4  \\ \midrule
        Less than once a week & 1   \\ \bottomrule
    \end{tabularx}
    \caption{Overview of the self-reported frequency of English usage of the non-native annotators.}
    \label{tab:enfreq}
\end{table}

\subsection{Prompt length}
Table~\ref{tab:prompt_len} shows the average prompt length per dataset and per group.
It is interesting to note the large difference for the CNN dailymail dataset, where the non-native English speakers have provided on average longer summaries. For the Western native English group versus the not Western native English group, the summaries for the latter are on average 10 words longer than for the former.

\begin{table}[]
\scriptsize
\centering
\resizebox{!}{1.9cm}{\begin{tabularx}{0.5\textwidth}{l!{\vrule}ll!{\vrule}ll}
\toprule
\multirow{2}{*}{\makecell{Dataset ids}} & \multicolumn{2}{l}{Native or not} & \multicolumn{2}{l}{\makecell{Western native or not}}   \\ \cmidrule(lr){2-3} \cmidrule(lr){4-5}
   & \makecell{native} & \makecell{non-native} & \makecell{not Western native} & \makecell{Western native}  \\ \midrule
\textbf{0} & 11.08  & 10.17& 11.52&  10.14  \\
\textbf{1} & 9.15& 8.94 & 8.31 & 9.26    \\
\textbf{2} & 9.40& 9.71 & 9.73 & 9.58    \\
\textbf{3} & 14.95& 13.00 & 14.8 & 13.39  \\
\textbf{4} & 7.56& 7.57  & 7.41& 7.61   \\
\textbf{5} & 7.53& 7.74  & 6.91 & 7.93   \\
\textbf{6} & 59.32& 66.14 & 56.48 & 66.41   \\
\textbf{7} & 12.09& 11.74 & 12.04 & 11.77  \\
\textbf{8} & 11.28& 11.38 & 11.38 & 11.34   \\
\textbf{9} & 25.91& 28.30  & 24.65 & 28.66   \\ \bottomrule
\end{tabularx}}
\caption{Average prompt length per group and dataset.}
\label{tab:prompt_len}
\end{table}

\hfill\break

\subsection{Time analysis annotators}
Table~\ref{tab:ann_dur_min} shows an overview of the average duration of annotating one example per group in minutes. Table~\ref{tab:ann_dur_h}, on the other hand, shows the average time for annotating the given set in hours.
\begin{table}[]
    \centering
    \begin{tabularx}{0.45\textwidth}{l!{\vrule}X}
    \toprule
        \textbf{group} & \textbf{duration (in min)} \\ \midrule
        native & 2.07 \\
        non native & 3.27\\
        Western native & 1.87\\
        not Western native & 3.25\\ \bottomrule
    \end{tabularx}
    \caption{Average duration of annotating 1 example per group, in minutes}
    \label{tab:ann_dur_min}
\end{table}

\begin{table}[ht]
    \scriptsize
    \centering
    \resizebox{!}{1.9cm}{\begin{tabularx}{0.5\textwidth}{l!{\vrule}ll!{\vrule}ll}
    \toprule
        Set-ids & \multicolumn{2}{l}{Native or not} & \multicolumn{2}{l}{\makecell{Western native or not}} \\ \cmidrule(lr){2-3} \cmidrule(lr){4-5}
        & native & non-native & \makecell{Western native} & \makecell{not Western native}	 \\ \hline
        10 & 2.70 & 3.06 & 2.33 & 3.12 \\	
        20 & 3.22 & 4.01 & 3.72 & 3.71	 \\
	30 & 3.05 & 5.95 & 3.68 & 5.09	 \\
	40 & 3.31 & 5.12 & 2.95 & 5.04	 \\
	50 & 4.82 & 4.68 & 2.98 & 5.04	 \\
	60 & 1.30 & 3.37 & 1.33 & 3.10	 \\
	70 & 1.39 & 3.66 & 1.45 & 3.46	 \\
	80 & 2.39 & 4.54 & 2.39 & 4.54	 \\
	90 & 4.63 & 7.14 & 4.63 & 7.14	 \\
	100 & 2.79 & 3.39 & 2.03 & 3.65	 \\ \bottomrule
    \end{tabularx}}
    \caption{Average overall duration of annotating all annotations in set per group, in hours}
    \label{tab:ann_dur_h}
\end{table}

\section{Prompt templates}
\label{sec:prompts}
\noindent\textbf{General} \newline
system prompt:\textit{You are a helpful assistant.}

\noindent chat history: \newline
User: \textit{task definition},\newline
Assistant:\textit{'Understood'},\newline
User: \textit{Full English Prompt}\newline

\noindent\textbf{Add Native} \newline
system prompt:\textit{You are a helpful assistant. Respond as if you are interacting with a native English speaker}

\noindent chat history: \newline
User: \textit{[task definition]},\newline
Assistant:\textit{Understood},\newline
User: \textit{[Full English Prompt]}\newline

\noindent\textbf{Add Non-Native} \newline
system prompt:\textit{You are a helpful assistant. Respond as if you are interacting with a non-native English speaker}

\noindent chat history: \newline
User: \textit{[task definition]},\newline
Assistant:\textit{Understood},\newline
User: \textit{[Full English Prompt]}\newline

Since we found that some of the models were not following the task definitions correctly for some of the tasks, we added extra instructions as to how the model should reply. Table~\ref{tab:extra_task_def} shows the instructions that were added to the task definition for the different datasets.
\begin{table}[]
\small
    \centering
    \begin{tabularx}{0.45\textwidth}{l!{\vrule}X}
    \toprule
        \textbf{Task} & \textbf{Task definition}  \\ \midrule
        StoryCloze &  Only respond with the predicted last sentence. \\
        AmazonFood & Only respond with the rating.\\
        McTaco & Only respond with "yes" or "no". \\
        TweetQA & Only respond with "yes" or "no". \\
        CNN Dailymail & Only respond with the news article. \\
        CODA19 & Only respond with the paragraph. \\
        Paraphrase & Only respond with the paraphrased sentence. \\
        Commonsense & Only respond with the letter indicating the most corresponding reason. \\ \bottomrule
    \end{tabularx}
    \caption{Overview of the added instructions per dataset to ensure consistent answers from the LLMs.}
    \label{tab:extra_task_def}
\end{table}

\section{Checkpoints models and hyperparameters}
\label{sec:checkpoints}

We used the following checkpoints of the different models:

\noindent \textbf{GPT 3.5} was made by OpenAI\footnote{\url{https://openai.com/index/gpt-3-5-turbo-fine-tuning-and-api-updates/}}. We used \textit{gpt-3.5-turbo-0125}.

\noindent \textbf{GPT 4o} was made by OpenAI\footnote{\url{https://openai.com/index/hello-gpt-4o/}}. We used \textit{gpt-4o-2024-05-13}.

\noindent \textbf{Haiku} was made by Anthropic~\citep{anthropic2024claude}. We used \textit{claude-3-Haiku-20240307}.

\noindent \textbf{Sonnet} was made by Anthropic~\citep{anthropic2024claude}. We used \textit{claude-3-Sonnet-20240229}.

\noindent \textbf{Qwen 7B} is an open source model made by the Alibaba group~\citep{bai2023qwen}. We used \textit{Qwen/Qwen1.5-7B-Chat}

We have generated the responses for the classification tasks with a temperature set at 0 and for the generation tasks at 0.7

\section{Evaluation metrics}
\label{sec:eval}
All generative performance metrics were gathered using Llama-3.3-70B-Instruct. Below, the prompts for the different metrics were given. These are based on the ones used for the Summeval dataset by~\citet{Bavaresco2024JUDGE_BENCH}. Depending on the generative task, the prompts differ slightly. we use the following structure: 
Final prompt= ""start\_prompt + doc\_gen: [model generated reply] + metric""
where the start prompt for fluency is: 

\emph{'You will be given a doc\_gen generated based on a doc\_comp. Your task is to rate the doc\_gen on one metric. Please make sure you read and understand these instructions carefully. Please keep this document open while reviewing, and refer to it as needed.}

and the start prompt for the other metrics is:
\emph{You will be given a doc\_gen and a doc\_comp. Your task is to rate the doc\_gen on one metric. Please make sure you read and understand these instructions carefully. Please keep this document open while reviewing, and refer to it as needed.}

doc\_gen and doc\_comp depend on the dataset. They are a 'closing sentence' and a 'story' for Story Cloze; a 'news article' and a 'summary' for CNN Dailymail; an 'article' and a 'title' for CODA19; and a 'paraphrased sentence' and a 'sentence' for Paraphrase.

The metric is one of the following: 

\noindent \textbf{Fluency}:
\newline
\emph{ Evaluation Criteria: Fluency (1-3): the quality of the doc\_gen in terms of grammar, spelling, punctuation, word choice, and sentence structure. Assign a score on a scale of 1 to 3 where: - 1: Poor. The doc\_gen has many errors that make it hard to understand or sound unnatural.- 2: Fair. The doc\_gen has some errors that affect the clarity or smoothness of the text, but the main points are still comprehensible. - 3: Good. The doc\_gen has few or no errors and is easy to read and follow. Evaluation Form (scores ONLY):- Fluency:}

\noindent \textbf{Coherence.} \newline 
The definition depends on the dataset. 
For CNN Dailynews and CODA19 it is as follows:\newline
\emph{Evaluation Criteria: Coherence (1-5) - the collective quality of all sentences. We align this dimension with the DUC quality question of structure and coherence whereby the doc\_gen should be well-structured and well-organized. The doc\_gen should not just be a heap of related information, but should build from sentence to a coherent body of information about a topic. Evaluation Steps: 1. Read the doc\_comp carefully and identify the main topic and key points.2. Read the doc\_gen and compare it to the doc\_comp. Check if the doc\_gen covers the main topic and key points of the doc\_gen, and if it presents them in a clear and logical order.3. Assign a score for coherence on a scale of 1 to 5, where 1: Very low coherence ; 2: Low coherence; 3: Mediocre coherence ; 4: High coherence ; 5: Very high coherence. Evaluation Form (scores ONLY):- Coherence:}

For Paraphrase it is as follows:\newline
\emph{Evaluation Criteria: Coherence (1-5) - The overall quality of the paraphrased sentence in terms of logical flow, structure, and alignment with the original sentence. A coherent paraphrase should preserve the meaning of the original sentence, avoid redundancy, and introduce variation without altering the main idea. The paraphrased sentence should not feel disjointed or incomplete but should read smoothly as a standalone sentence. Evaluation Steps: 1. Read the doc\_comp carefully and identify the main topic and key points.  2. Read the doc\_gen and compare it to the doc\_comp.  3. Assign a score for coherence on a scale of 1 to 5, where 1: Very low coherence ; 2: Low coherence; 3: Mediocre coherence ; 4: High coherence ; 5: Very high coherence. Evaluation Form (scores ONLY): - Coherence:}

For Story Cloze it is as follows:
\emph{Evaluation Criteria: Coherence (1-5) - the collective quality of all sentences. We align this dimension with the DUC quality question of structure and coherence whereby the sentences should be well-structured and well-organized. The sentences should not just be a heap of related information, but should build from sentence to a coherent story.Evaluation Steps: 1. Read the doc\_comp carefully and identify the main topic and key points. 2. Read the doc\_gen and compare it to the doc\_comp. Check if the sentences are clear and in a logical order.\ 3. Assign a score for coherence on a scale of 1 to 5, where 1: Very low coherence ; 2: Low coherence; 3: Mediocre coherence ; 4: High coherence ; 5: Very high coherence.  Evaluation Form (scores ONLY): - Coherence:}

\noindent \textbf{Relevance.} \newline 
The definition depends on the dataset. 
For Story Cloze it is as follows: \newline
\emph{Evaluation Criteria: Relevance (1-5) - The degree to which the generated doc\_gen effectively reflects the main themes and purpose of the doc\_comp. A relevant closing sentence should provide a meaningful and appropriate conclusion, aligning with the tone and key points of the narrative.  Evaluation Steps: 1. Read the doc\_comp and the doc\_gen carefully. 2. Compare the doc\_gen to the doc\_comp and identify the main points of the doc\_comp. 3. Assess how well the doc\_gen concludes the doc\_comp, and how much irrelevant or redundant information it contains. 4. Assign a relevance score from 1 to 5 where 1: Very low relevance ; 2: Low relevance; 3: Mediocre relevance ; 4: High relevance ; 5: Very high relevance.   Evaluation Form (scores ONLY): - Relevance:}

For all other datasets it is as follows: \newline
\emph{Evaluation Criteria: Relevance (1-5) - inclusion of important content from the doc\_comp. The doc\_gen should include all important information from the doc\_comp.  Evaluation Steps: 1. Read the doc\_comp and the doc\_gen carefully. 2. Compare the doc\_gen to the doc\_comp and identify the main points of the doc\_comp. 3. Assess how well the doc\_gen covers the main points of the doc\_comp, and how much irrelevant or redundant information it contains. 4. Assign a relevance score from 1 to 5 where 1: Very low relevance ; 2: Low relevance; 3: Mediocre relevance ; 4: High relevance ; 5: Very high relevance.  Evaluation Form (scores ONLY): - Relevance:}

\section{Distribution Amazon food reviews}
\label{app:amazon}
\begin{figure*}[ht]
\centering
    \begin{subfigure}[]{0.45\textwidth}
    \includegraphics[width=\textwidth]{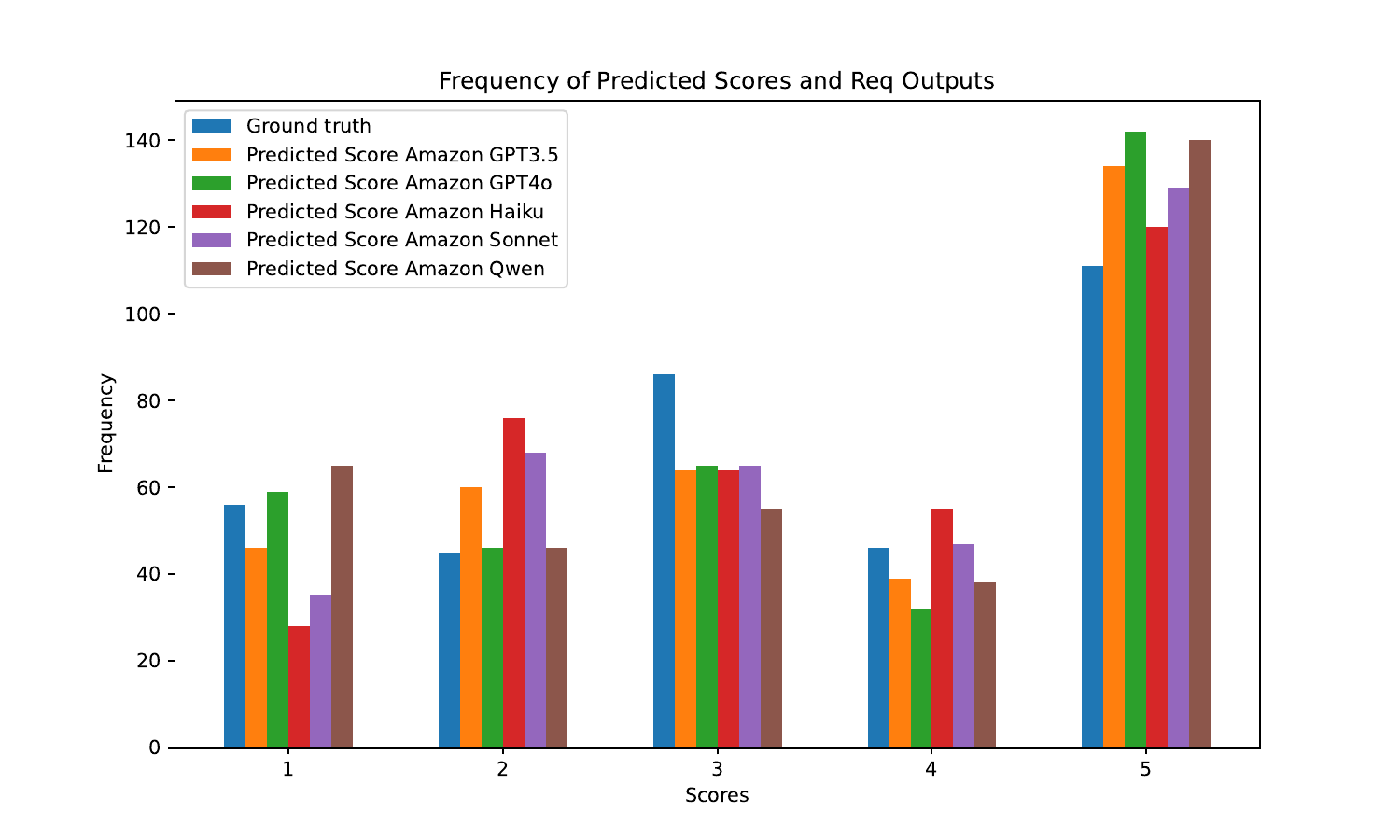}
    \caption{Overview of the predictions for the Western native English speakers.}
    \label{fig:amazon_west}
    \end{subfigure}%
    \hfill
    \begin{subfigure}[]{0.45\textwidth}
    \centering
    \includegraphics[width=\textwidth]{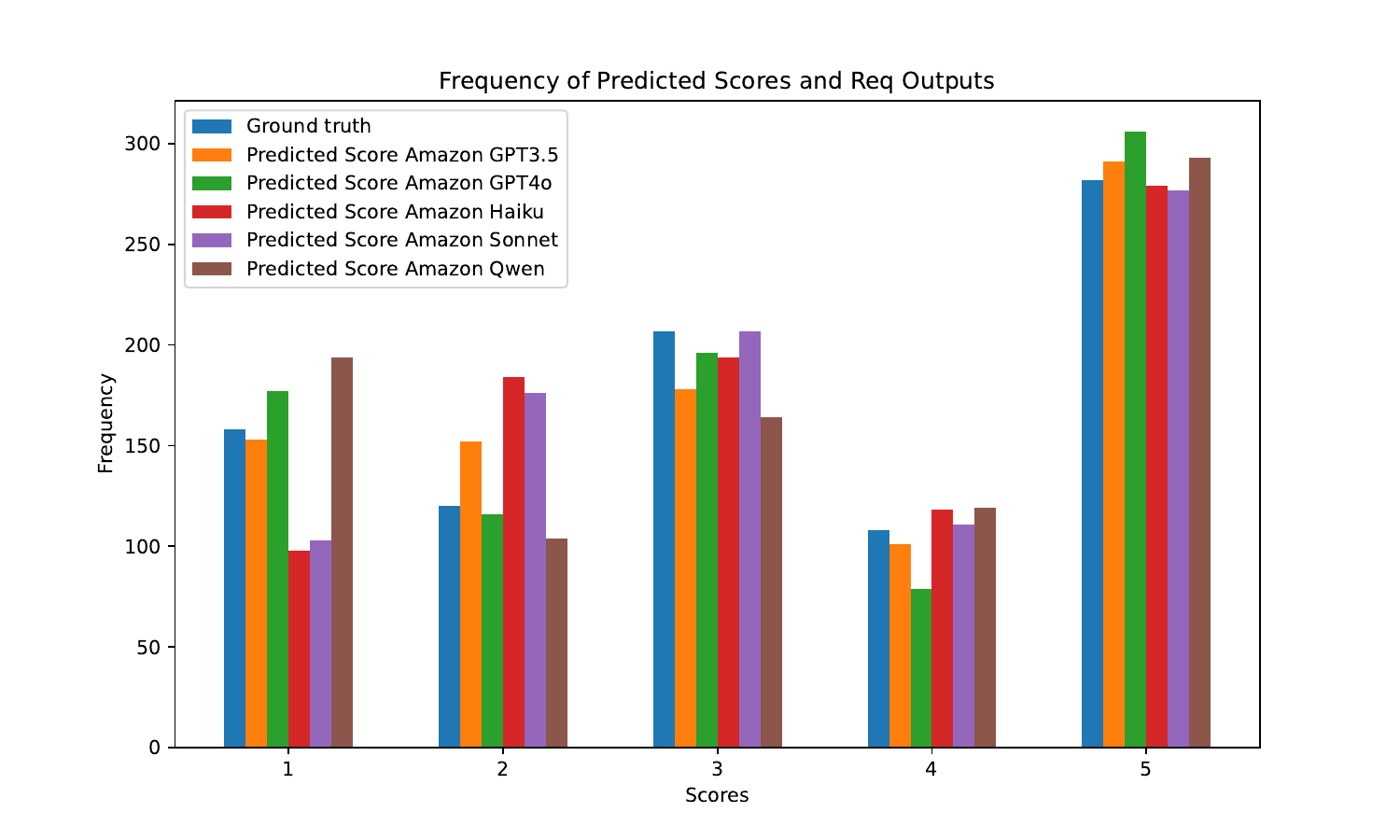}
            \caption{Overview of the predictions for the non-native English speakers.}
            \label{fig:amazon_non_native}
    \end{subfigure}
    \vfill
    \begin{subfigure}[]{0.45\textwidth}
    \includegraphics[width=\textwidth]{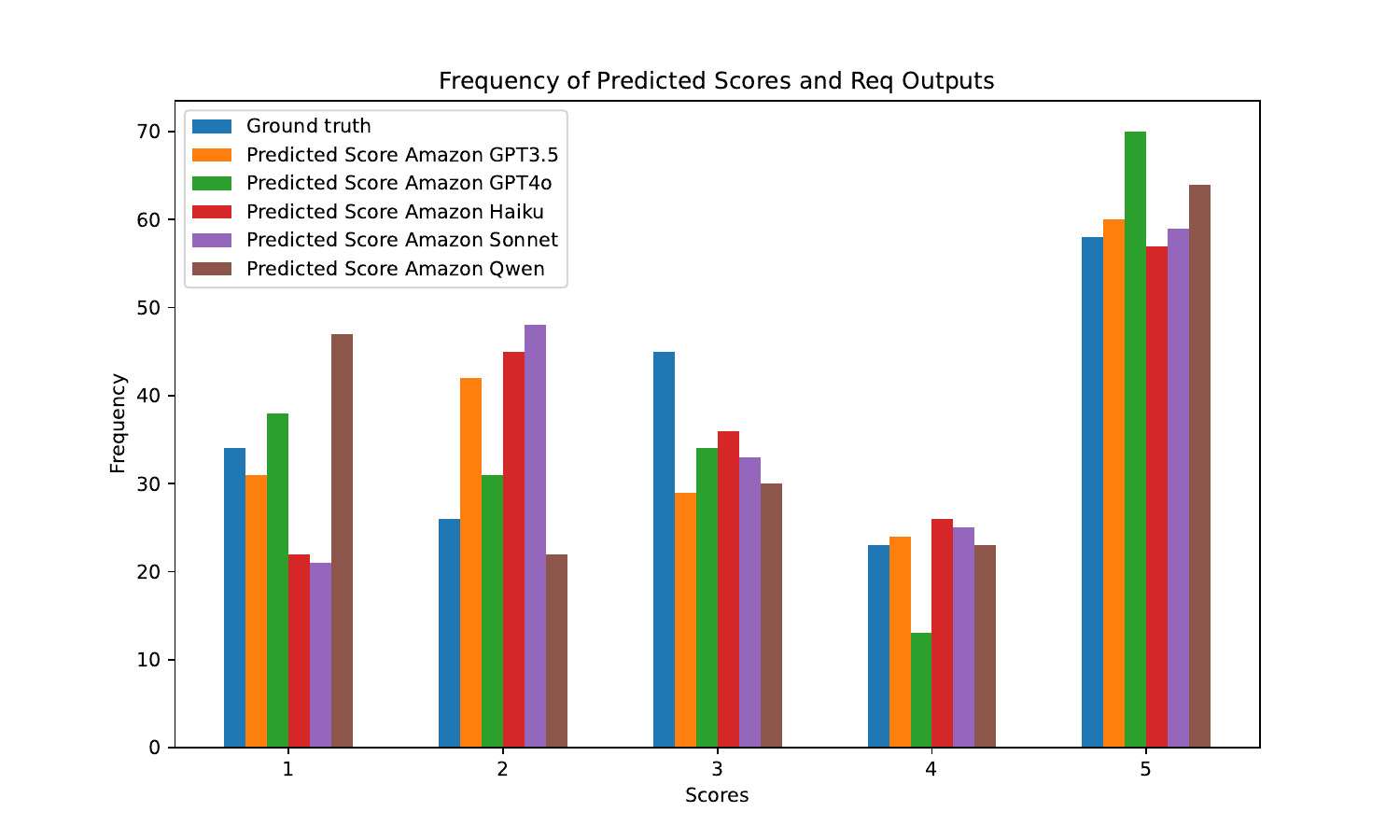}
        \caption{Overview of the predictions for the native English speakers that are not western native.}
        \label{fig:amazon_not_western}
    \end{subfigure}
    \caption{Overall classifications for Western native, native that are not Western native, and non-native English speakers}\label{fig:amazon_nat_non}
\end{figure*}

Figure~\ref{fig:amazon_nat_non} shows an overview of the wrong predictions of the AmazonFood review dataset for the different groups and models for one of the three runs. This shows the distribution between what was predicted and what should be predicted. We only consider here the cases where the model predicted one of the given ratings, and excluded cases where no prediction was given. As shown, for both the native and Western native group, we find a large amount of misclassification for the highest rating. Additionally, neutral is not often predicted for these classes compared to the other groups.

\section{Results Sonnet different languages}
When adding that the model is interacting with a non-native English speaker, we find that Sonnet starts to answer in different languages.
We find that for 668 prompts the model answers in Spanish, for 25 sentences in French, and for 5 sentences in Indonesian. There were a couple of other languages that also occurred sporadically. An overview is shown in Table~\ref{tab:other_languages_in_Sonnet}. However, these answers were not related to the native language of the prompt writer. This phenomenon was encountered mainly for the Timetravel dataset. Interestingly, this effect was not seen for the other models, not even for Haiku. 
\begin{table}[]
    \centering
    \begin{tabularx}{0.35\textwidth}{l!{\vrule}l}
        \toprule
        \textbf{Language} & \textbf{Times Occurring} \\ 
        \midrule
         es& 668 \\
         fr & 25 \\
         id & 5 \\
         it & 2 \\
         lt & 1 \\
         sw & 1 \\
         ru & 1 \\
         \bottomrule
    \end{tabularx}
    \caption{Occurrences of different languages in Sonnet}
    \label{tab:other_languages_in_Sonnet}
\end{table}

\section{Example Paraphrase}
\label{app:paraphrase}
As said, there are differences between native and non-native speakers as to how they perceived the paraphrasing task.
For example given this desired output: \textit{At this time of rapid change, those who lag behind fall into irrelevance}. Native speakers came up with very freely paraphrased sentences, such as: \textit{If you are not adapting to the quick changes of the world, you will not succeed.} while non-native speakers stuck to \textit{In this fast changing ages, whoever is lagging becomes irrelevant}. When giving these different sentences to the model to paraphrase, the result for the more freely paraphrased sentences might cause the model to shift away further from the initial sentence or gold answer.

\section{Classification results} \label{app:class_results_overall_per_model}
\begin{figure*}
    \centering
    \includegraphics[width=\linewidth]{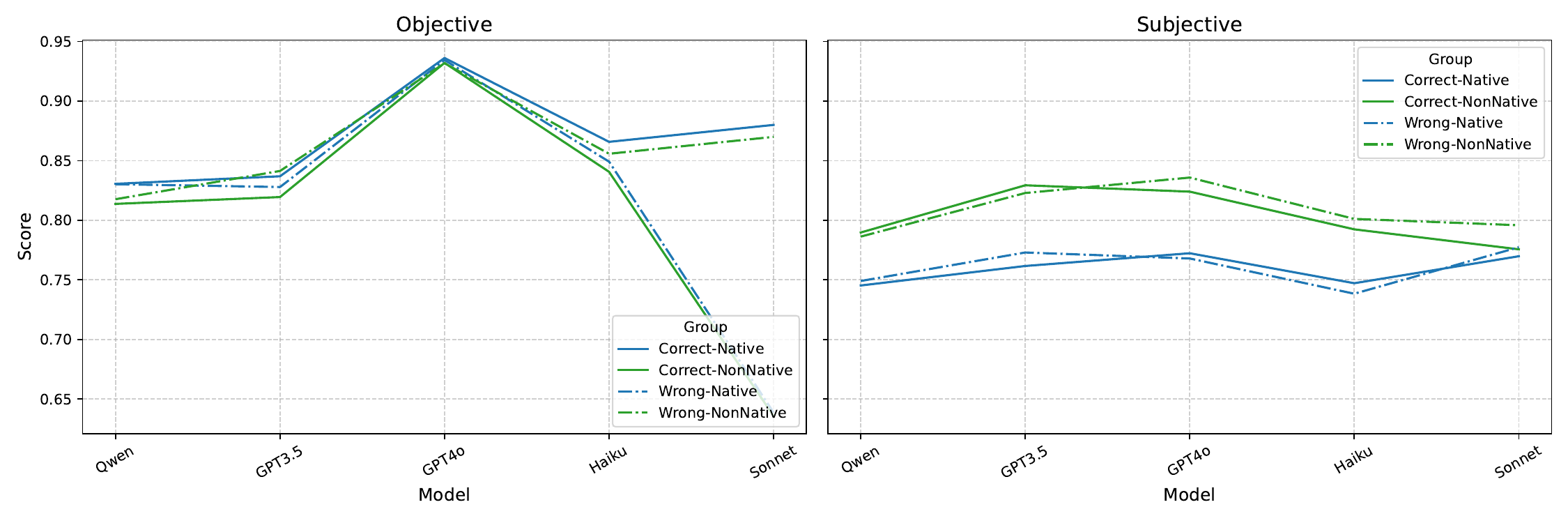}
    \caption{Classification results per model and classification task when information about the nativeness of the prompt writer was added. We clearly see how Sonnet is highly influenced by this additional information.}
    \label{fig:class_results_per_model_nativeness}
\end{figure*}

Figure~\ref{fig:class_results_per_model_nativeness} shows the accuracy scores for the objective and subjective classification tasks per model when information about the nativeness of the prompt writer is added. We see how sonnet clearly performs differently than the other models.

\section{LLM as a judge: Generative results} \label{app:gen_results_overall_per_model}
\begin{figure*}
    \centering
    \includegraphics[width=\linewidth]{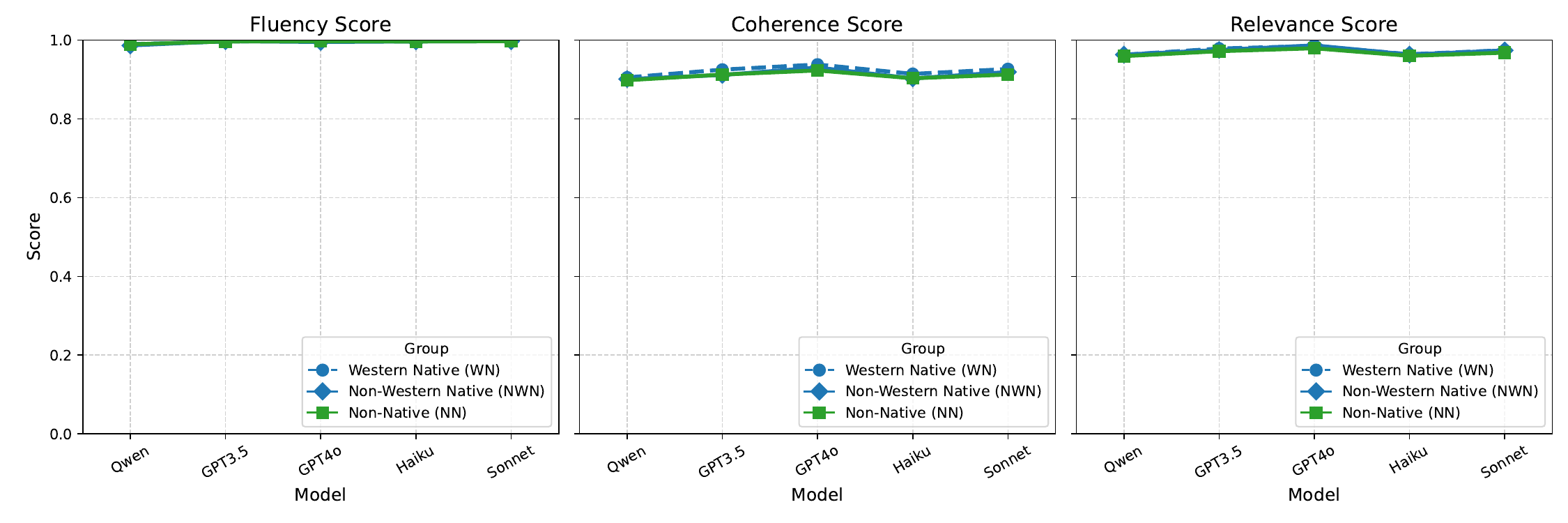}
    \caption{This Figure shows the performance of (western) native speakers and non-native speakers. We see how the highest performance for Coherence and is obtained for the western native group across all different models. The relevance scores show slightly less difference between groups, but the non-native and not western native group performs worse overall. The fluency scores are similar for all groups. We rescaled the results so that they range from 0 to 1.}
    \label{fig:gen_results_per_model}
\end{figure*}

\begin{figure*}[h!]
    \centering
    \includegraphics[width=\linewidth]{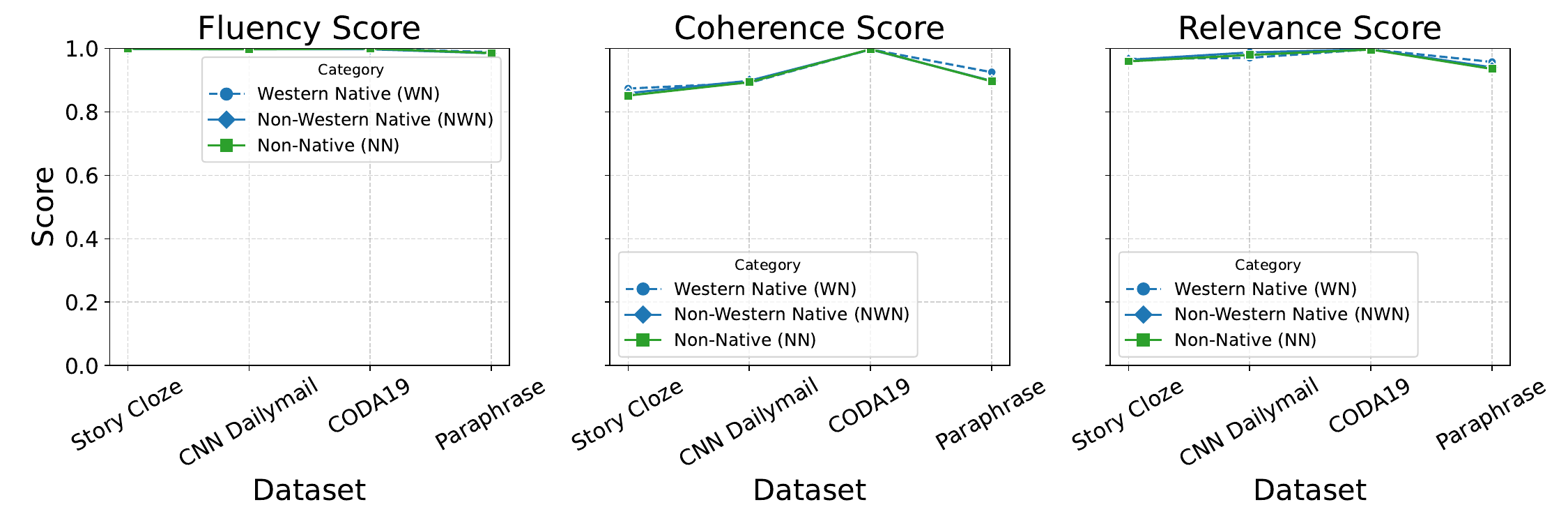}
    \caption{This figure shows the overall performance across the three groups: (western) native speakers and non-native speakers. However, when looking into the coherence metric, we do see a preference for the western native group. The results show how there is no difference regarding fluency and only a slight performance difference when comparing the native categories with the non-native category for relevance. }
    \label{fig:gen_results_per_dataset}
\end{figure*}

Figure~\ref{fig:gen_results_per_model} shows how similar behavior is found across all three performance metrics per model. Moreover, Figure~\ref{fig:gen_results_per_dataset} shows the results per dataset for the generative results.

\section{Additional Analysis}
In this section, we include some extra analyses on the performance of the different groups within the non-native English speakers. More specifically, we add the results per level of English proficiency, as well as per frequency of English. We see that there are differences in performance across the different groups. 
\subsection{Classification results}
\begin{figure}
    \centering
    \includegraphics[width=\linewidth]{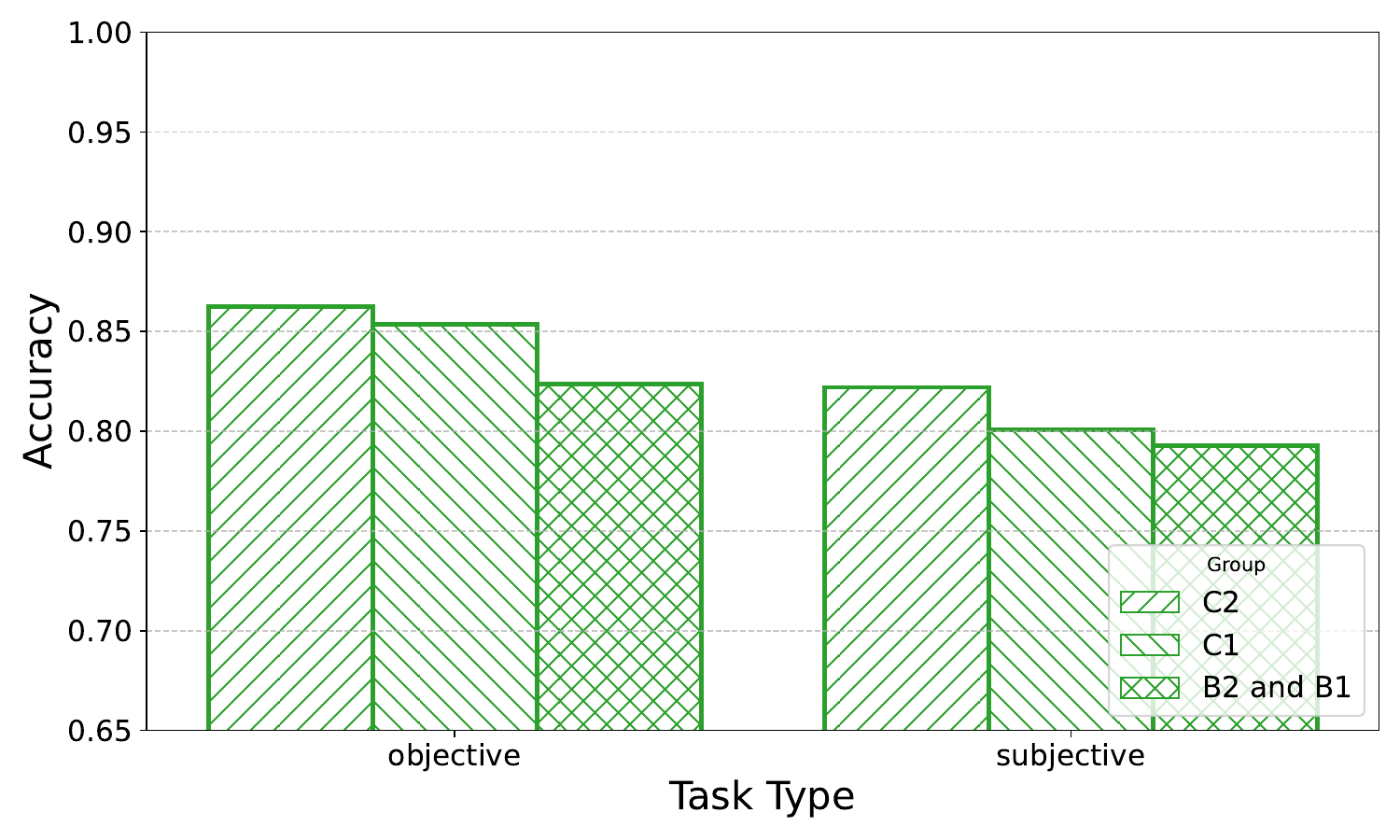}
    \caption{This figure shows the performance of English non-native speakers per self-reported level of English for the classification tasks.We adjusted the y-axis to range from 0.65 to 1 for clarity. }
    \label{fig:level_class}
\end{figure}

\begin{figure}
    \centering
    \includegraphics[width=\linewidth]{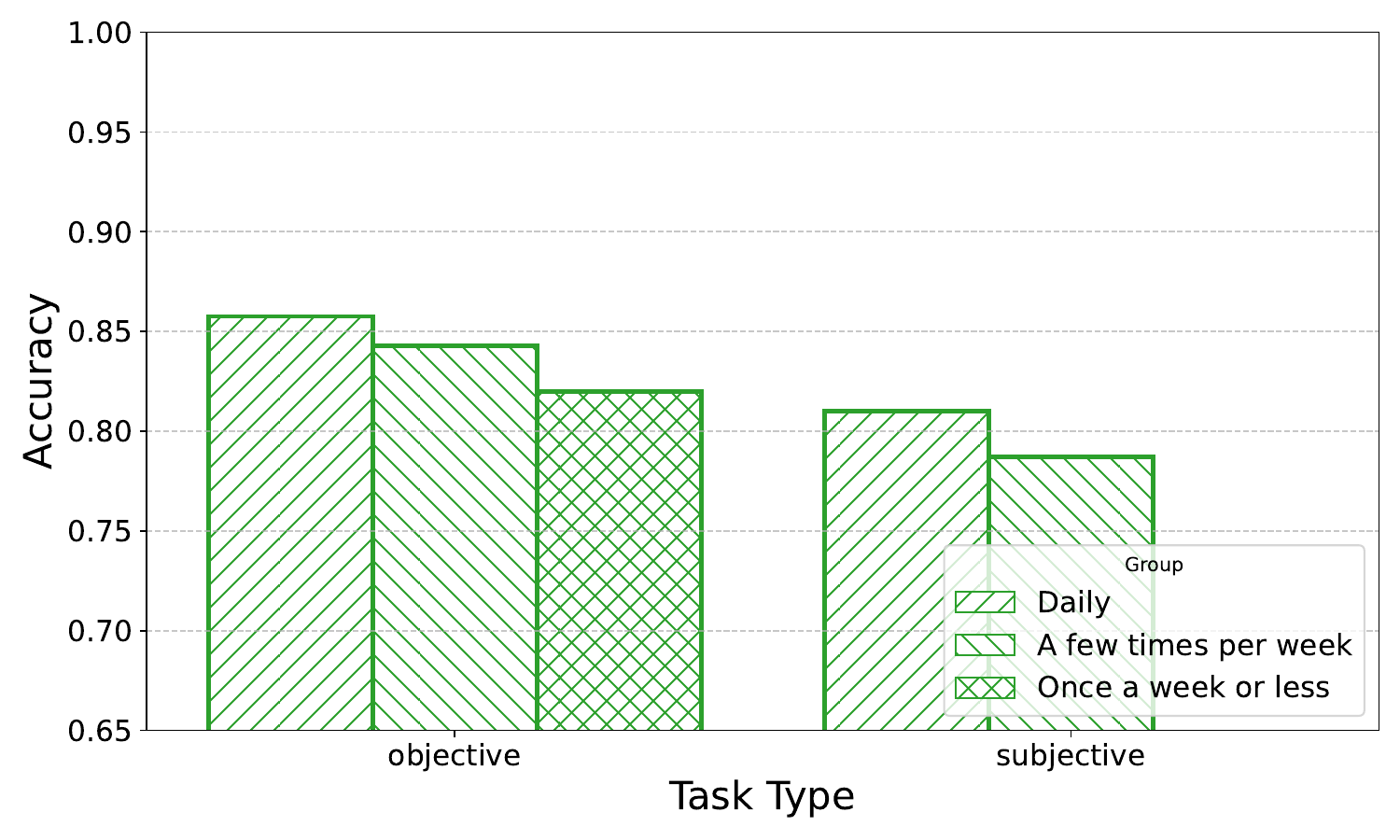}
    \caption{This figure shows the performance of English non-native speakers per self-reported frequency of English usage for the classification tasks. We adjusted the y-axis to range from 0.65 to 1 for clarity.}
    \label{fig:freq_class}
\end{figure}

For the classification results, we see a clear connection between performance and level of English, and frequency of usage of English. The groups with the highest levels of English also obtain better results. This is shown in Figures \ref{fig:level_class} and \ref{fig:freq_class}.

\begin{figure}
    \centering
    \includegraphics[width=\linewidth]{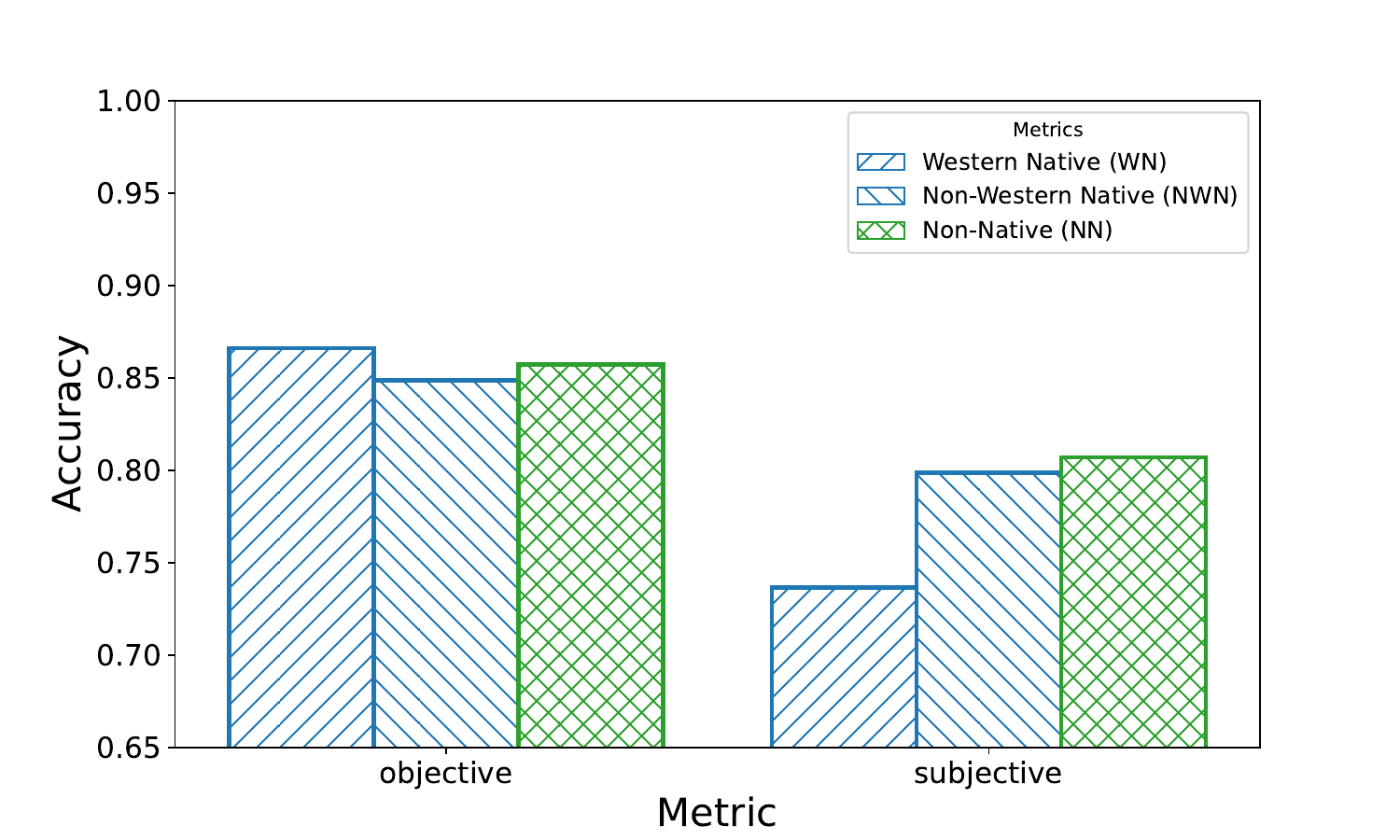}
    \caption{This figure shows the performance of the three groups only including C2 and C1 level English speakers. We adjusted the y-axis to range from 0.65 to 1 for clarity.}
    \label{fig:C1C2_class_comp}
\end{figure}

As we saw a performance difference, in terms of levels of English, we also compare the results when only taking into account level C1 and C2 non-native English speakers. The results are shown in Figure~\ref{fig:C1C2_class_comp}. Here, we still see the same order in performance as in Figure~\ref{fig:class_results_normal_prompt} was shown. However, now there is a clearer performance difference between the natives that are not western native and the non-native group. 

\subsection{Generative Results}
\begin{figure}
    \centering
    \includegraphics[width=\linewidth]{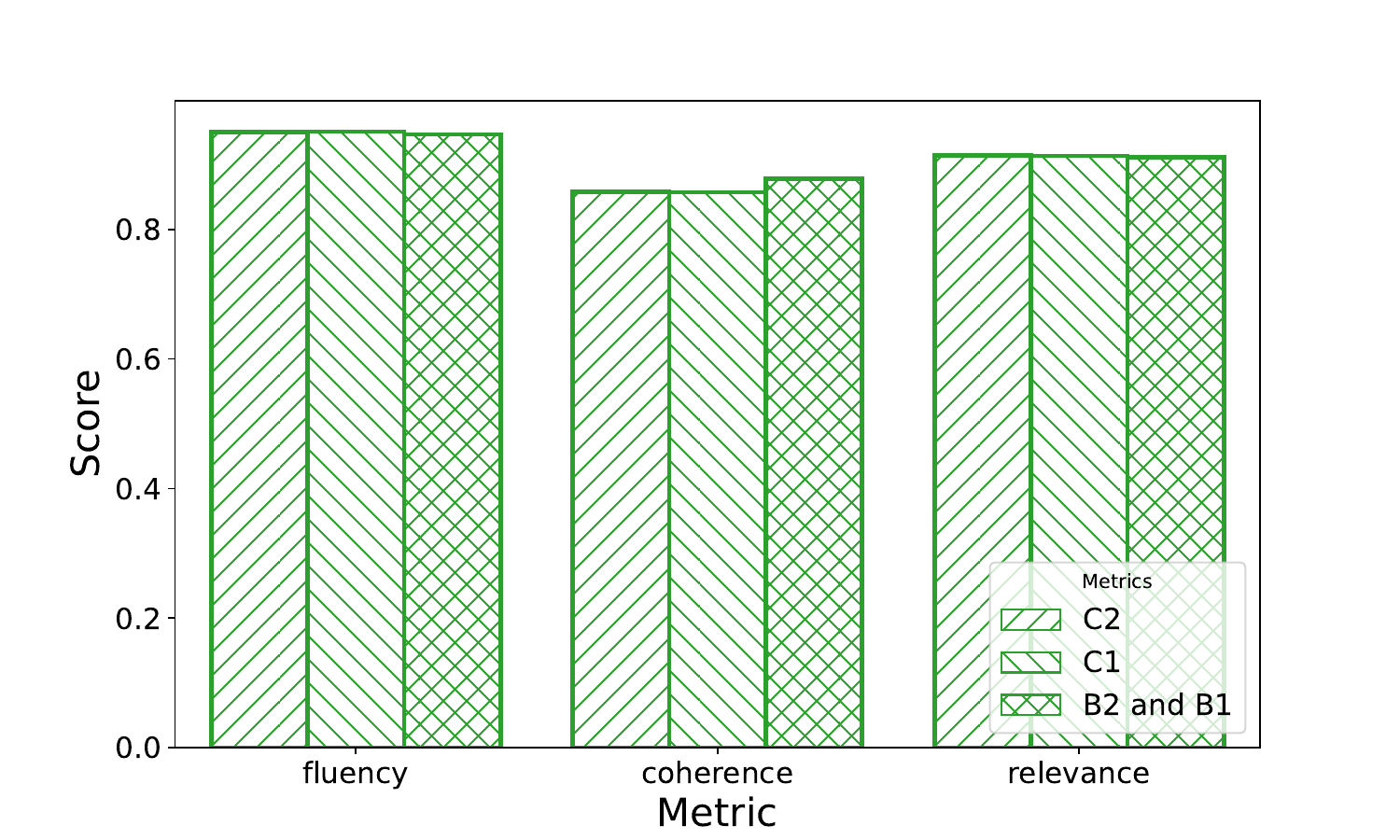}
    \caption{This figure shows the performance of English non-native speakers per self-reported level of English for the generative tasks. We rescaled the results so that they range from 0 to 1.}
    \label{fig:level_en_gen}
\end{figure}

\begin{figure}
    \centering
    \includegraphics[width=\linewidth]{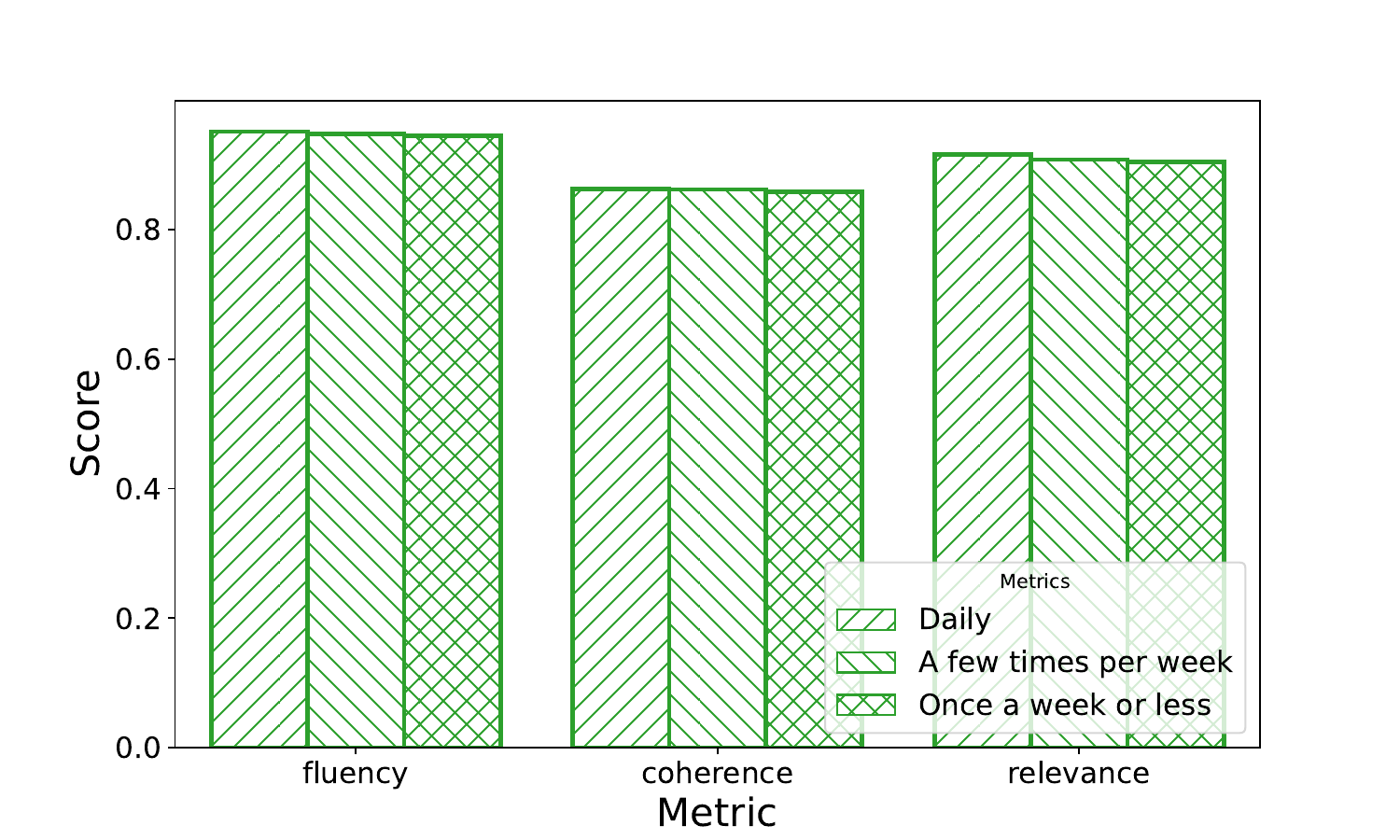}
    \caption{This figure shows the performance of English non-native speakers per self-reported frequency of English usage for the generative tasks. We rescaled the results so that they range from 0 to 1.}
    \label{fig:freq_gen}
\end{figure}

For the generative tasks, however, we do not see clear differences in terms of frequency of English usage and performance, as shown in~\ref{fig:level_en_gen} and \ref{fig:freq_gen}. Only the people with the lowest level of English proficiency perform better in terms of coherence, which is unexpected.

\begin{figure}
    \centering
    \includegraphics[width=\linewidth]{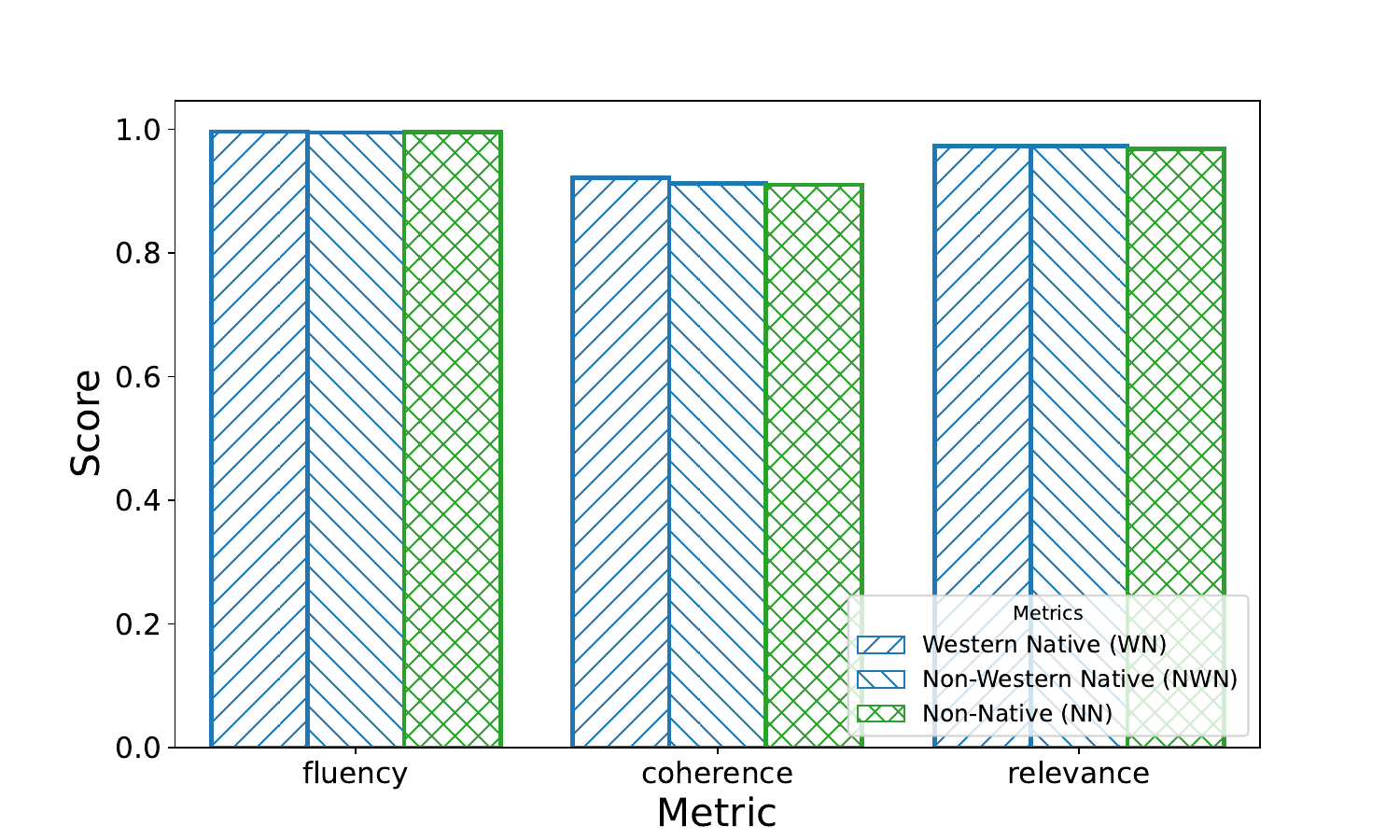}
    \caption{This figure shows the generative results only for the C2 and C1-level speakers per group.We rescaled the results so that they range from 0 to 1.}
    \label{fig:gen_Results_onlyC2_andC1}
\end{figure}

When analyzing the performance differences only for the groups with highest proficiency (C2 and C1), as shown in Figure~\ref{fig:gen_Results_onlyC2_andC1}, we see similar findings to Figure~\ref{fig:gen_results}.

\section{Robustness analysis annotations}
The dataset used in this paper is designed to be parallel, ensuring the same base samples for different annotator groups. We divided the dataset into multiple sets as shown in Table~\ref{tab:annotators_per_group}. As shown, the WN and NN groups have annotated all existing sets. The NWN group annotated most of the dataset, but did not annotate two out of the ten sets. Our dataset was constructed through random sampling and annotators from the three groups annotated the majority of sets. Below we provide a robustness check on the overlapping subsets. 

\begin{table}[]
    \centering
    \footnotesize
    \begin{tabular}{c|cc}
    \toprule
    \textbf{Full dataset} & & \\
    \toprule
        Group & \makecell{Objective \\ Classification} & \makecell{Subjective \\ Classification}  \\ \midrule
        WN & 0.8661 & 0.7366 \\
        NWN & 0.8487 & 0.7986 \\
        NN & 0.8518 & 0.8039 \\ \bottomrule
    \textbf{Only Overlapping Sets} & & \\ \toprule
    Group & \makecell{Objective \\ Classification} & \makecell{Subjective \\ Classification}   \\ \midrule
    WN & 0.8655 & 0.7366 \\
    NWN & 0.8487 & 0.7986 \\
    NN & 0.8492 & 0.8040 \\
         \bottomrule
    \end{tabular}
    \caption{Performance for the classification tasks on the full dataset and only considering overlapping samples.}
    \label{tab:robustness}
\end{table}

Table~\ref{tab:robustness} shows how the results remain consistent when analyzing the full dataset and only the overlapping sets for both the objective and subjective classification tasks. This illustrates how our findings demonstrate genuine performance differences rather than artifacts of different datasets, confirming that the dataset's structure does not impact the findings.

\end{document}